\documentclass[10pt]{article}

\usepackage{graphicx}
\usepackage{color}

\usepackage[authoryear]{natbib}

\usepackage{bm}
\usepackage{amsmath}
\usepackage{amsfonts}
\usepackage{amsthm}
\usepackage{amssymb}
\usepackage{mathtools}
\mathtoolsset{showonlyrefs=true}
\newtheorem{theorem}{Theorem}
\newtheorem{lemma}{Lemma}

\usepackage{url}
\usepackage{subcaption}
\usepackage{algorithm}
\usepackage{algorithmic}

\makeatletter
\renewcommand{\ALG@name}{Procedure}
\makeatother

\usepackage[hidelinks]{hyperref}

\title{\vspace{-10mm}Statistical Test for Attention Map in\\Vision Transformer}

\date{\today}

\makeatletter
\def\@fnsymbol#1{\ensuremath{\ifcase#1\or
{1}\or
{2}\or
{3}\or
{4}\or
{5}\or
{\dagger}\or
\else\@ctrerr\fi}}
\makeatother

\author{
Tomohiro Shiraishi\thanks{Nagoya University} ,
Daiki Miwa\thanks{Nagoya Institute of Technology} ,\\
Teruyuki Katsuoka\footnotemark[1] ,
Vo Nguyen Le Duy\thanks{University of Information Technology, Ho Chi Minh City, Viet Nam} \thanks{Vietnam National University, Ho Chi Minh City, Viet Nam} \thanks{RIKEN} ,\\
Kouichi Taji\footnotemark[1] ,
Ichiro Takeuchi\footnotemark[1] \footnotemark[5] \thanks{Corresponding author. e-mail: ichiro.takeuchi@mae.nagoya-u.ac.jp}
}
\begin{document}

\maketitle

\thispagestyle{empty}

\begin{abstract}
    \noindent
    The Vision Transformer (ViT) demonstrates exceptional performance in various computer vision tasks.
Attention is crucial for ViT to capture complex wide-ranging relationships among image patches, allowing the model to weigh the importance of image patches and aiding our understanding of the decision-making process.
However, when utilizing the attention of ViT as evidence in high-stakes decision-making tasks such as medical diagnostics, a challenge arises due to the potential of attention mechanisms erroneously focusing on irrelevant regions.
In this study, we propose a statistical test for ViT's attentions, enabling us to use the attentions as reliable quantitative evidence indicators for ViT's decision-making with a rigorously controlled error rate.
%
%
%
Using the framework called selective inference, we quantify the statistical significance of attentions in the form of $p$-values, which enables the theoretically grounded quantification of the false positive detection probability of attentions.
%
%
We demonstrate the validity and the effectiveness of the proposed method through numerical experiments and applications to brain image diagnoses.

\end{abstract}

\newpage
\section{Introduction}
\label{sec:introduction}
The Vision Transformer (ViT)~\citep{dosovitskiy2020image} demonstrates exceptional performance across a spectrum of computer vision tasks by replacing traditional Convolutional Neural Networks (CNNs) with transformer-based architectures.
ViT divides images into fixed-size patches and processes them using self-attention mechanisms, capturing wide-range dependencies.
This enables the model to effectively learn spatial relationships and contextual information, surpassing the limitations of CNNs in handling global context~\citep{wu2020visual,henaff2020data,xiao2021early,touvron2021training,jia2021scaling,khan2022transformers}.

In ViT, attention plays a pivotal role in capturing complex visual relationships by allowing the model to weigh the importance of different image regions.
The interpretability of attention mechanisms is crucial for understanding how the model makes decisions.
ViT's attention mechanisms enable the identification of salient features and contribute to the model's ability to recognize patterns.

However, when utilizing the attention of ViT as evidence in high-stakes decision-making tasks such as medical diagnostics or autonomous driving~\citep{dai2021transmed,he2023transformers,prakash2021multi,hu2022toward}, a challenge arises due to the potential of attention mechanisms erroneously focusing on irrelevant regions.
In this study, we propose a statistical test for ViT's attentions, enabling the quantification of the false positive (FP) detection probability of attentions in the form of $p$-values.
This enables us to use the attentions as reliable quantitative evidence indicators for ViT's decision-making with a rigorously controlled error rate.

To our knowledge, there are no prior studies that investigate the statistical significance of ViT's attentions.
The challenge in assessing the statistical significance of ViT's attentions stems from the inherent selection bias in ViT's attention mechanism.
Testing image patches with high attention is biased, given that the ViT selects these patches by looking at the image itself.
Consequently, it becomes imperative to develop an appropriate statistical test that can account for selection bias by properly considering the complex attention mechanism of ViT.

In this study, we address this challenge by employing selective inference (SI)~\citep{lee2016exact,taylor2015statistical}.
SI is a statistical inference framework that has gained recent interest for testing data-driven hypotheses.
By considering the selection process itself as part of the statistical analysis, SI effectively addresses the selection bias issue in statistical testing when the hypotheses are selected in a data-driven manner.

SI was initially developed for the statistical inference of feature selection in linear models~\citep{fithian2015selective,tibshirani2016exact,loftus2014significance,suzumura2017selective,le2021parametric,sugiyama2021more} and later extended to other problem settings~\citep{lee2015evaluating,choi2017selecting,chen2020valid,tanizaki2020computing,duy2020computing,gao2022selective}.
In the context of deep learning, SI was first introduced by \citep{duy2022quantifying} and~\citep{miwa2023valid}, where the authors proposed a computational algorithm for SI by exploiting the fact that a class of CNNs can be described as piecewise linear functions of the input image.
In this study, we introduce a new computational approach to develop SI for ViT's attentions, as they do not exhibit such piecewise linearity.

\begin{figure}[htbp]
    \begin{minipage}[b]{1.0\linewidth}
        \centering
        \includegraphics[width=0.9\linewidth]{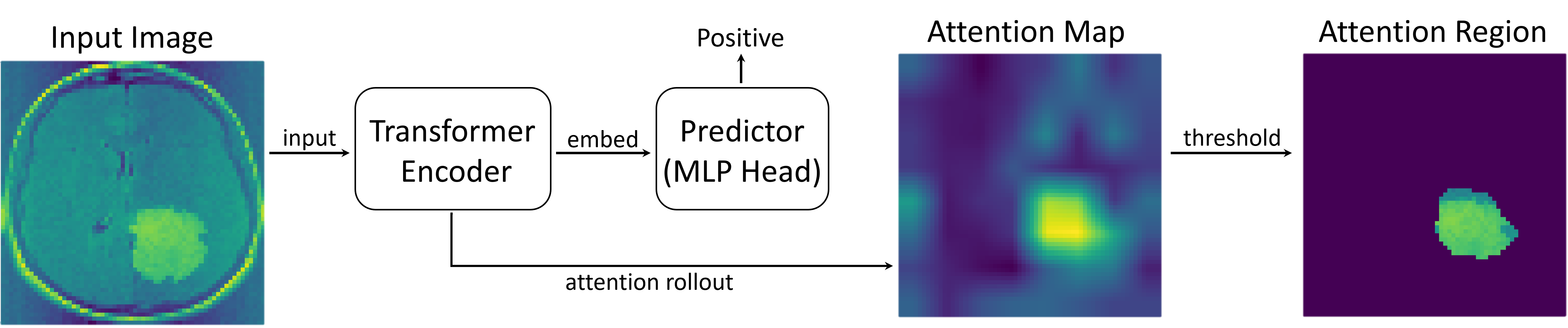}
        \subcaption{
            Brain image with tumor. The naive $p$-value is 0.000 (true positive) and the selective $p$-value is 0.000 (true positive).
        }
    \end{minipage}
    \begin{minipage}[b]{1.0\linewidth}
        \centering
        \includegraphics[width=0.9\linewidth]{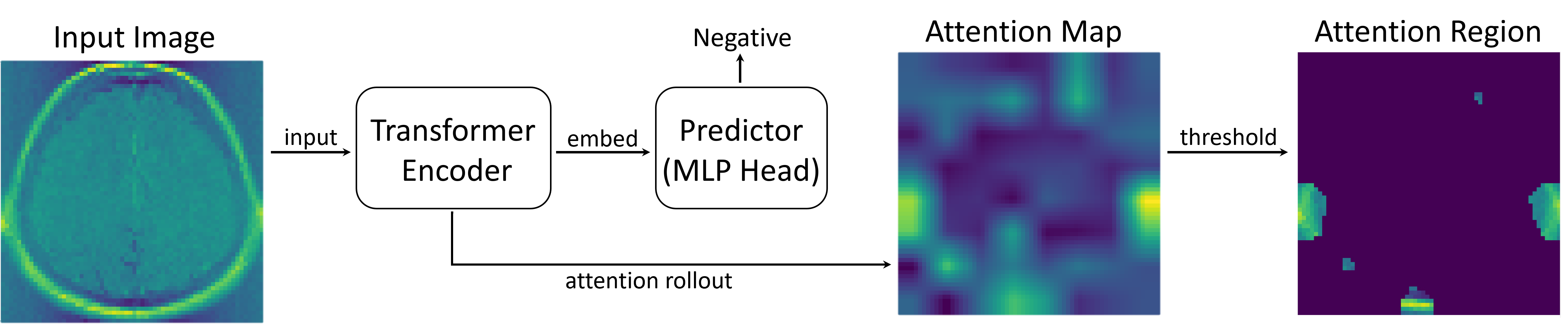}
        \subcaption{
            Brain image without tumor. The naive $p$-value is 0.000 (false positive) and the selective $p$-value is 0.801 (true negative).
        }
    \end{minipage}
    \caption{
        Schematic illustration of the problem setup and the proposed method on a brain image dataset.
        By inputting a brain image into the trained ViT classifier, the attention map is obtained, which indicates the area on which the ViT model focuses.
        Our objective is to provide the statistical significance of the attention map using the $p$-value.
        To achieve this objective, we consider testing the attention region, which consists of pixels with high attention levels by thresholding the attention map.
        The results suggest that the naive $p$-value (see \S\ref{sec:sec4}) cannot be used to properly control the false positive (type I error) rate.
        Instead, the selective $p$-value (introduced in \S\ref{sec:sec2}) can be used to detect true positives while controlling the false positive rate at the specified level.
    }
    \label{fig:schematic_illustration_problem_setup}
\end{figure}

Figure~\ref{fig:schematic_illustration_problem_setup} illustrates the problem setup considered in this study, where we applied a naive statistical test, which does not consider selection bias, and our proposed statistical test to brain image diagnosis task.
The upper panel shows a brain image with a tumor region, in which we want the attentions to be declared as statistically significant (with a small $p$-value).
Here, both the naive test and the proposed test conclude that the identified attention is statistically significant with $p$-values nearly 0.
In contrast, the lower panel displays a brain image without tumor regions, in which we want the attentions to be determined as statistically not significant (with a large $p$-value).
In this case, the naive test falsely detects significance (false positive) with an almost zero $p$-value, while the proposed method yields a $p$-value of 0.801, concluding that it is not statistically significant (true negative).

Our contributions in this study are as follows.
The first contribution is the introduction of a theoretically guaranteed framework for testing the statistical significance of ViT's Attention (\S\ref{sec:sec2}).
The second contribution involves the development of the SI method for ViT's attention, for which we introduce a new computational method for computing the $p$-values without selection bias (\S\ref{sec:sec3}).
The third contribution involves demonstrating the effectiveness of the proposed method through its applications to synthetic data simulations and brain image diagnosis (\S\ref{sec:sec4}).
For reproducibility, our implementation is available at \url{https://github.com/shirara1016/statistical_test_for_vit_attention}.
%
%
%
%
%
%

\newpage
\section{Statistical Test for ViT's Attentions}
\label{sec:sec2}
In this study, we aim to quantify the statistical significance of the attention regions identified by a trained ViT model.
The details of the structure of the ViT model we used are shown in Appendix~\ref{app:vit_structure}.
\paragraph{Notations.}
Let us consider an $n$-dimensional image as a random variable
\begin{equation}
    \bm{X}=
    (X_1,\ldots,X_n)= \bm{\mu} +\bm{\epsilon},\
    \bm{\epsilon}\sim \mathcal{N}(\bm{0}, \Sigma),
\end{equation}
where $\bm{\mu}\in\mathbb{R}^n$ is the pixel intensity vector and $\bm{\epsilon}\in\mathbb{R}^n$ is the noise vector with covariance matrix $\Sigma\in\mathbb{R}^{n\times n}$.
We do not pose any assumption on the true pixel intensity $\bm{\mu}$, while we assume that the noise vector $\bm{\epsilon}$ follows the Gaussian distribution with the covariance matrix $\Sigma$ known or estimable from external independent data\footnote{
    We discuss the robustness of the proposed method when the covariance matrix is unknown and the noise deviates from the Gaussian distribution in Appendix~\ref{app:robustness}
}.
We define the computation of the attention map as a mapping $\mathcal{A}\colon \mathbb{R}^n\ni \bm{X}\mapsto \mathcal{A}(\bm{X})\in[0,1]^n$, which takes an image $\bm{X}$ as input and outputs attention scores $\mathcal{A}_i(\bm{X})\in[0,1]$ for each pixel $i\in[n]$.
The details of the attention map computation are given in Appendix~\ref{app:attention_map}.
We define the attention region $\mathcal{M}_{\bm{X}}$ of an image $\bm{X}$ as the set of pixels with attention scores greater than a given threshold value $\tau\in(0,1)$, i.e.,
\begin{equation}
    \label{eq:attention_region}
    \mathcal{M}_{\bm{X}}=\{i\in[n]\mid \mathcal{A}_i(\bm{X})>\tau\}.
\end{equation}
\paragraph{Statistical Inference.}
To quantify the statistical significance of the attention region $\mathcal{M}_{\bm{X}}$ of an image $\bm{X}$, we propose to consider the following hypothesis testing problem:
\begin{equation}
    \label{eq:hypothesis_testing}
    \begin{gathered}
        \mathrm{H}_{0}\colon
        \frac{1}{|\mathcal{M}_{\bm{X}}|}
        \sum_{i\in\mathcal{M}_{\bm{X}}}\mu_i
        =
        \frac{1}{|\mathcal{M}_{\bm{X}}^c|}
        \sum_{i\notin\mathcal{M}_{\bm{X}}}\mu_i\\
        \text{v.s.}\\
        \mathrm{H}_{1}\colon
        \frac{1}{|\mathcal{M}_{\bm{X}}|}
        \sum_{i\in\mathcal{M}_{\bm{X}}}\mu_i
        \neq
        \frac{1}{|\mathcal{M}_{\bm{X}}^c|}
        \sum_{i\notin\mathcal{M}_{\bm{X}}}\mu_i,
    \end{gathered}
\end{equation}
where $\mathrm{H}_0$ is the null hypothesis that the mean pixel intensity inside and outside the attention region are equal, while $\mathrm{H}_1$ is the alternative hypothesis that they are not equal.
A reasonable choice of the test statistic for the statistical test in~\eqref{eq:hypothesis_testing} is the difference in the average pixel values between inside and outside the attention region, i.e.,
\begin{equation}
    \bm{\eta}_{\mathcal{M}_{\bm{X}}}^\top \bm{X}
    =
    \frac{1}{|\mathcal{M}_{\bm{X}}|}
    \sum_{i\in\mathcal{M}_{\bm{X}}}\mu_i
    -
    \frac{1}{|\mathcal{M}_{\bm{X}}^c|}
    \sum_{i\notin\mathcal{M}_{\bm{X}}}\mu_i,
\end{equation}
where $\bm{\eta}_{\mathcal{M}_{\bm{X}}}= \frac{1}{|\mathcal{M}_{\bm{X}}|}\bm{1}_{\mathcal{M}_{\bm{X}}}^n-\frac{1}{|\mathcal{M}_{\bm{X}}^c|}\bm{1}_{\mathcal{M}_{\bm{X}}^c}^n$ is a vector that depends on the attention region $\mathcal{M}_{\bm{X}}$, and $\bm{1}_{\mathcal{C}}^n\in\mathbb{R}^n$ is an $n$-dimensional vector whose elements are set to $1$ if they belong to the set $\mathcal{C}\subset[n]$, and $0$ otherwise.
In this study, we consider the following standardized test statistic:
\begin{equation}
    T(\bm{X})
    =
    \frac
    {\bm{\eta}_{\mathcal{M}_{\bm{X}}}^\top \bm{X}}
    {\sqrt{\bm{\eta}_{\mathcal{M}_{\bm{X}}}^\top\Sigma\bm{\eta}_{\mathcal{M}_{\bm{X}}}}}.
\end{equation}
The $p$-value for the hypothesis testing problem in~\eqref{eq:hypothesis_testing} can be used to quantify the statistical significance of the attention region $\mathcal{M}_{\bm{X}}$.
Given a significance level $\alpha\in(0,1)$ (e.g., 0.05), we reject the null hypothesis $\mathrm{H}_{0}$ if the $p$-value is less than $\alpha$, indicating that the attention region $\mathcal{M}_{\bm{X}}$ is significantly different from the outside of the attention region.
Otherwise, we fail to state that the attention region $\mathcal{M}_{\bm{X}}$ is statistically significant.

Our main idea in this formulation is to quantify whether pixels selected as attention regions by ViT are statistically significantly different from the regions that were not selected.
Although we consider the average difference in pixel values in the above formulation for simplicity, similar formulations are also possible for other image features obtained by applying appropriate image filters.

\paragraph{Conditional Distribution.}
To compute the $p$-value, we need to identify the sampling distribution of the test statistic $T(\bm{X})$.
However, as the vector $\bm{\eta}_{\mathcal{M}_{\bm{X}}}$ depends on the attention region $\mathcal{M}_{\bm{X}}$ (i.e., depends on $\bm{X}$ through a complicated computation in the ViT model), the sampling distribution of the test statistic $T(\bm{X})$ is too complicated to characterize.
Then, we consider the conditional sampling distribution of the test statistic $T(\bm{X})$ given the event $\{\mathcal{M}_{\bm{X}}=\mathcal{M}_{\bm{X}^\mathrm{obs}}\}$, i.e.,
%
\begin{equation}
    \label{eq:cond_dist}
    T(\bm{X})\mid \{\mathcal{M}_{\bm{X}}=\mathcal{M}_{\bm{X}^\mathrm{obs}}\},
\end{equation}
where $\bm{X}^\mathrm{obs}$ represents the observed image $\bm{X}$.
This conditioning means that we consider the rarity of the observation $\bm{X}^\mathrm{obs}$ only in the case where the same attention region $\mathcal{M}_{\bm{X}}$ as observed $\mathcal{M}_{\bm{X}^\mathrm{obs}}$ is obtained.
The advantage of considering the conditional sampling distribution in~\eqref{eq:cond_dist} is that, by conditioning on the attention region $\mathcal{M}_{\bm{X}}$, the test statistic $T(\bm{X})$ is written as a linear function of $\bm{X}$, which allows us to characterize the sampling property of the test statistic $T(\bm{X})$.
\paragraph{Selective $p$-value.}
Statistical hypothesis testing based on the conditional sampling distribution has been studied within the framework of SI (also known as post-selection inference).
In this study, we also utilize the SI framework to perform statistical hypothesis testing in~\eqref{eq:hypothesis_testing} based on the conditional sampling distribution in~\eqref{eq:cond_dist}.
For the tractable computation of the conditional sampling distribution in~\eqref{eq:cond_dist}, as is done in other SI studies we consider an additional condition on the sufficient statistic of the nuisance parameter $\mathcal{Q}_{\bm{X}}$, defined as
\begin{equation}
    \mathcal{Q}_{\bm{X}}
    =
    \left(
    I_n
    -
    \frac
    {\Sigma\bm{\eta}_{\mathcal{M}_{\bm{X}}}\bm{\eta}_{\mathcal{M}_{\bm{X}}}^\top}
    {\bm{\eta}_{\mathcal{M}_{\bm{X}}}^\top\Sigma\bm{\eta}_{\mathcal{M}_{\bm{X}}}}
    \right)
    \bm{X}.
\end{equation}
The selective $p$-value is then computed as
\begin{equation}
    \label{eq:selective_pvalue}
    p_\mathrm{selective} = \mathbb{P}_{\mathrm{H}_0}
    (
    |T(\bm{X})| > |T(\bm{X}^\mathrm{obs})|
    \mid
    \bm{X}\in\mathcal{X}
    ),
\end{equation}
where $\mathcal{X}=\{
    \bm{X}\in\mathbb{R}^n\mid
    \mathcal{M}_{\bm{X}}=\mathcal{M}_{\bm{X}^\mathrm{obs}},
    \mathcal{Q}_{\bm{X}}=\mathcal{Q}_{\bm{X}^\mathrm{obs}}\}$.
Based on the statistical theory developed in SI literature~\citep{fithian2014optimal}, it is possible to show the following theorem for the selective $p$-value.
%
\begin{theorem}
    \label{thm:property_of_pvalue}
    Under the null hypothesis $\mathrm{H}_0$ in \eqref{eq:hypothesis_testing}, for any $\alpha\in(0,1)$, the selective $p$-value in~\eqref{eq:selective_pvalue} satisfies the following property
    \begin{equation}
        \label{eq:property_of_pvalue}
        \mathbb{P}_{\mathrm{H}_0}
        \left(
        p_\mathrm{selective}\leq \alpha
        \mid
        \mathcal{M}_{\bm{X}}=\mathcal{M}_{\bm{X}^\mathrm{obs}}
        \right)
        =
        \alpha. 
    \end{equation}
\end{theorem}
This theorem guarantees that the selective $p$-value is uniformly distributed under the null hypothesis $\mathrm{H}_0$ and then used to conduct the valid statistical inference for the attention region $\mathcal{M}_{\bm{X}}$.
Here, it is important to note that the selective $p$-value is dependent on the $\mathcal{Q}_{\bm{X}}$, but the property in~\eqref{eq:property_of_pvalue} is satisfied without this additional condition, because we can marginalize over all the values of $\mathcal{Q}_{\bm{X}}$.
This theorem can be shown similarly to Theorem 5.2 in \citep{lee2016exact}.
%

\newpage
\section{Computing Selective $p$-values}
\label{sec:sec3}
In this section, we propose a novel computation procedure for computing the selective $p$-values in \eqref{eq:selective_pvalue}.
\paragraph{Characterization of the Conditional Data Space.}
To compute the selective $p$-value in~\eqref{eq:selective_pvalue}, we need to characterize the conditional data space $\mathcal{X}$.
According to the conditioning on the nuisance parameter $\mathcal{Q}_{\bm{X}}=\mathcal{Q}_{\bm{X}^\mathrm{obs}}$, the conditional data space $\mathcal{X}$ is restricted to a one-dimensional line in $\mathbb{R}^n$.
Therefore, the set $\mathcal{X}$ can be re-written, using a scalar parameter $z\in\mathbb{R}$, as
\begin{equation}
    \mathcal{X}=
    \{
    \bm{X}(z)\in\mathbb{R}^n\mid \bm{X}(z)=\bm{a}+\bm{b}z,\ z\in\mathcal{Z}
    \}
\end{equation}
where vectors $\bm{a},\bm{b}\in\mathbb{R}^n$ are defined as
\begin{equation}
    \bm{a}=\mathcal{Q}_{\bm{X}^\mathrm{obs}},\
    \bm{b}=
    \left.
    \Sigma\bm{\eta}_{\mathcal{M}_{\bm{X}^\mathrm{obs}}}
    \middle/
    \sqrt{
    \bm{\eta}_{\mathcal{M}_{\bm{X}^\mathrm{obs}}}^\top
    \Sigma \bm{\eta}_{\mathcal{M}_{\bm{X}^\mathrm{obs}}}
    }
    \right.,
\end{equation}
and the region $\mathcal{Z}$ is defined as
\begin{equation}
    \mathcal{Z} =
    \{
    z\in\mathbb{R}\mid
    \mathcal{M}_{\bm{a}+\bm{b}z}=\mathcal{M}_{\bm{X}^\mathrm{obs}}
    \}.
\end{equation}
This characterization of the conditional data space is first proposed in \citep{liu2018more} and used in many other SI studies.
Let us consider a random variable $Z\in\mathbb{R}$ and its observation $z^\mathrm{obs}\in\mathbb{R}$ such that they respectively satisfy $\bm{X} = \bm{a} + \bm{b}Z$ and $\bm{X}^\mathrm{obs} = \bm{a} + \bm{b}z^\mathrm{obs}$.
The selective $p$-value in~\eqref{eq:selective_pvalue} is re-written as
\begin{equation}
    \label{eq:selective_pvalue_one_dim}
    p_\mathrm{selective}
    =
    \mathbb{P}_{\mathrm{H}_0}
    (
    |Z|> |z^\mathrm{obs}|
    \mid
    Z\in\mathcal{Z}
    ).
\end{equation}
Because the unconditional variable $Z\sim \mathcal{N}(0,1)$ under the null hypothesis $\mathrm{H}_0$\nobreak
\footnote{
    The random variable $Z$ corresponds to the test statistic $T(\bm{X})$.
    Then, $Z\sim\mathcal{N}(0,1)$ is obtained by the linearity of the test statistic $T(\bm{X})$ with respect to $\bm{X}$ and the test statistic $T(\bm{X})$ is already standardized in the definition.
},
the conditional random variable $Z\mid Z\in\mathcal{Z}$ follows the truncated standard Gaussian distribution.
Once the truncated region $\mathcal{Z}$ is identified, the selective $p$-value in~\eqref{eq:selective_pvalue_one_dim} can be easily computed.
Thus, the remaining task is reduced to the characterization of $\mathcal{Z}$.
%
%
\paragraph{Reformulation of the Truncated Region.}
Based on the definition of the attention region in~\eqref{eq:attention_region}, the condition part of the set $\mathcal{Z}$ can be reformulated as
\begin{align}
                    &
    \mathcal{M}_{\bm{a}+ \bm{b}z}=\mathcal{M}_{\bm{X}^\mathrm{obs}} \\
    \Leftrightarrow &
    \{i\in[n]\mid \mathcal{A}_i(\bm{a}+ \bm{b}z) > \tau\}
    =
    \mathcal{M}_{\bm{X}^\mathrm{obs}}                               \\
    \Leftrightarrow &
    \begin{cases}
        \mathcal{A}_i(\bm{a}+ \bm{b}z) > \tau,\
        \forall i\in\mathcal{M}_{\bm{X}^\mathrm{obs}} \\
        \mathcal{A}_i(\bm{a}+ \bm{b}z) < \tau,\
        \forall i\notin\mathcal{M}_{\bm{X}^\mathrm{obs}}
    \end{cases}                 \\
    \Leftrightarrow &
    f_i(z) < 0,\ \forall i\in[n],
\end{align}
where $f_i\colon\mathbb{R}\to\mathbb{R},\ i\in[n]$ is defined as
\begin{equation}
    \label{eq:definition_of_fi}
    f_i(z) =
    \begin{cases}
        \tau - \mathcal{A}_i(\bm{a}+\bm{b}z) &
        (i\in\mathcal{M}_{\bm{X}^\mathrm{obs}}) \\
        \mathcal{A}_i(\bm{a}+\bm{b}z) - \tau &
        (i\notin\mathcal{M}_{\bm{X}^\mathrm{obs}})
    \end{cases}.
\end{equation}
Therefore, we can reformulate $\mathcal{Z}$ as
\begin{equation}
    \label{eq:reformulated_truncated_region}
    \mathcal{Z}=
    \bigcap_{i\in [n]}
    \{
    z\in\mathbb{R}\mid f_i(z) < 0
    \}.
\end{equation}
\paragraph{Selective $p$-value Computation by Adaptive Grid Search.}
The problem of finding $\mathcal{Z}$ in~\eqref{eq:reformulated_truncated_region} is reduced to the problem of enumerating all solutions to the nonlinear equations $f_i(z) = 0$ for each $i \in [n]$ in~\eqref{eq:definition_of_fi}.
The difficulty of this problem depends on the continuity, differentiability, and smoothness of the functions $f_i,\ i \in [n]$.
Fortunately, since the function $f_i$ is a part of the attention map computation in the ViT model, it is continuous, (sub)differentiable, and possesses a certain level of smoothness (except for pathological cases).
Assuming the certain degree of smoothness of the function $f_i$, by adaptively generating grid points in the one-dimensional space $z\in\mathbb{R}$ and computing the values of $f_i(z)$ at each grid point, it is possible to identify $\mathcal{Z}$ in~\eqref{eq:reformulated_truncated_region} with sufficient accuracy.
This further means that it is possible to compute the selective $p$-value in~\eqref{eq:selective_pvalue_one_dim} with sufficient accuracy (as stated in Theorem~\ref{thm:error_bound} later).
%
%
The overall procedure for estimating the selective $p$-value by an adaptive grid search method is summarized in Procedure~\ref{alg:adaptive_grid}.
Here, $S$ represents the grid search interval $[-S, S]$, $\varepsilon_\mathrm{min}$ and $\varepsilon_\mathrm{max}$ represent the minimum and maximum grid width, respectively.
Note that, in line 9 of Procedure~\ref{alg:adaptive_grid}, we added the interval $J(z^\mathrm{obs})$ that overlaps with the grid points for computational simplicity.
The key of Procedure~\ref{alg:adaptive_grid} lies in how to determine the adaptive grid size $d(z_j)$.
\begin{algorithm}
    \caption{Selective $p$-value Computation by Adaptive Grid Search}
    \label{alg:adaptive_grid}
    \begin{algorithmic}[1]
        \REQUIRE $S$, $\varepsilon_\mathrm{min}$, $\varepsilon_\mathrm{max}$, $\{f_i\}_{i\in[n]}$ and $z^\mathrm{obs}:=T(\bm{X}^\mathrm{obs})$
        \STATE $j\leftarrow 0, z_0\leftarrow -S$
        \WHILE{$z_j<S$}
        \STATE compute the adaptive grid width $d(z_j)$
        \STATE $z_{j+1}\leftarrow z_j + \min(\varepsilon_\mathrm{max},\max(d(z_j),\varepsilon_\mathrm{min}))$
        \STATE $j\leftarrow j+1$
        \ENDWHILE
        \STATE $d^\mathrm{obs}\leftarrow \min(\varepsilon_\mathrm{max}, d(z^\mathrm{obs}))$
        \STATE $J(z^\mathrm{obs})\leftarrow [z^\mathrm{obs}-d^\mathrm{obs}, z^\mathrm{obs}+d^\mathrm{obs}]$
        \STATE $\mathcal{Z}^\mathrm{grid}\leftarrow \cup_{j\mid z_j\in\mathcal{Z}}[z_j,z_{j+1}]\cup J(z^\mathrm{obs})$
        \STATE $p_\mathrm{grid}\leftarrow
            \mathbb{P}_{\mathrm{H}_0}(|Z|> |z^\mathrm{obs}|\mid Z\in\mathcal{Z}_\mathrm{grid})$, where $Z\sim\mathcal{N}(0,1)$
        \ENSURE $p_\mathrm{grid}$
    \end{algorithmic}
\end{algorithm}

\begin{figure}[htbp]
    \centering
    \includegraphics[width=0.7\linewidth]{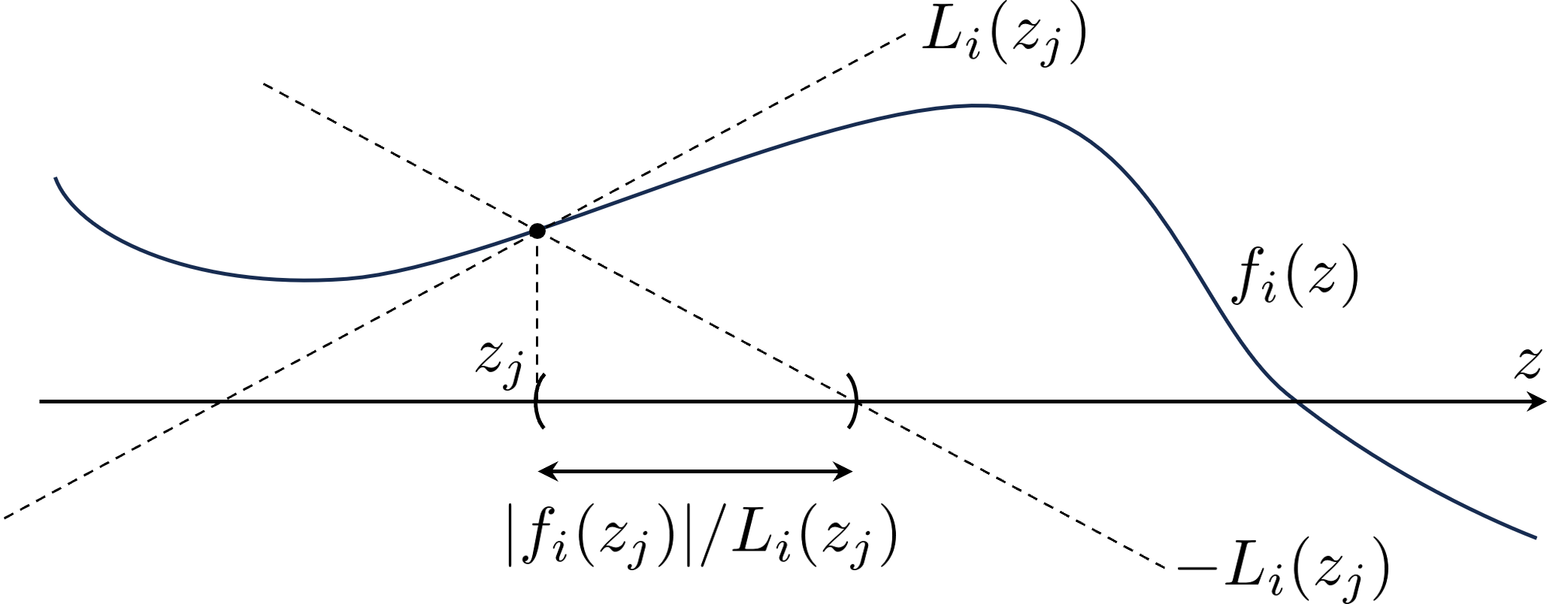}
    \caption{Schematic illustration of Lipschitz continuity.}
    \label{fig:lipschitz}
\end{figure}

\begin{figure}[htbp]
    \centering
    \includegraphics[width=1.0\linewidth]{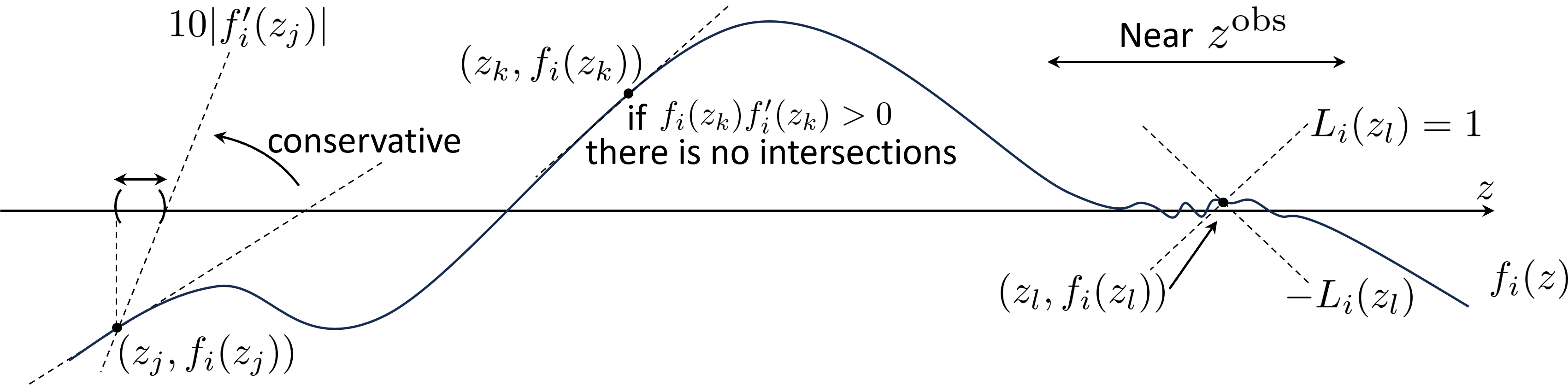}
    \caption{Schematic illustration of the introduced heuristics.}
    \label{fig:heuristics}
\end{figure}

%
%
The following theorem states that, by utilizing the Lipschitz constant of $f_i$, it is possible to appropriately determine the adaptive grid width $d(z_j)$ and compute the selective $p$-value with sufficient accuracy.
\begin{theorem}
    \label{thm:error_bound}
    Assume that for all $i\in[n]$, $f_i$ is differentiable and Lipschitz continuous.
    Assume further that for all $i\in[n]$, $f_i$ has at most only a finite number of zeros, at any of which the value of $f_i^\prime$ is non-zero.
    Define the grid width $d(z_j)$ as
    \begin{equation}
        d(z_j) =
        \begin{dcases}
            \min_{i\in[n],f_i(z_j)< 0}\frac{|f_i(z_j)|}{L_i(z_j)}    &
            (z_j\in\mathcal{Z})                                        \\
            \max_{i\in[n],f_i(z_j)\geq 0}\frac{|f_i(z_j)|}{L_i(z_j)} &
            (z_j\notin\mathcal{Z})
        \end{dcases},
    \end{equation}
    where $L_i(z_j)$ is the Lipschitz constant of $f_i$ in the $\varepsilon_\mathrm{max}$-neighborhood of $z_j$.
    Then, we have
    \begin{gather}
        |p_\mathrm{selective}-p_\mathrm{grid}|
        = O(\varepsilon_\mathrm{min}+\exp(-S^2/2)),\\
        \text{where $\varepsilon_\mathrm{min}\to 0, S\to\infty$}.
    \end{gather}
\end{theorem}
The proof of Theorem~\ref{thm:error_bound} is presented in Appendix~\ref{app:proof_of_thm}.
The following lemma suggests why it is reasonable to define the grid width as $d(z_j)$ in Theorem~\ref{thm:error_bound}.
\begin{lemma}
    \label{lem:grid_width}
    For the grid width $d(z_j)$ defined in Theorem~\ref{thm:error_bound}, we have
    \begin{align}
         & z_j\in\mathcal{Z}\Rightarrow
        [z_j, z_j+\min(\varepsilon_\mathrm{max}, d(z_j))]
        \subset \mathcal{Z},               \\
         & z_j\notin\mathcal{Z}\Rightarrow
        [z_j, z_j+\min(\varepsilon_\mathrm{max}, d(z_j))]
        \subset \mathbb{R}\setminus\mathcal{Z}.
    \end{align}
\end{lemma}
The proof of Lemma~\ref{lem:grid_width} is presented in Appendix~\ref{app:proof_of_lem}.
The main idea of the proof is that $f_i$ has the same sign on the interval $[z_j, z_j+\min(\varepsilon_\mathrm{max}, |f_i(z_j)|/L_i(z_j))]$ from Lipschitz continuity (as shown in Figure~\ref{fig:lipschitz}).
In Procedure~\ref{alg:adaptive_grid}, we take the $\max$ operation in line 4 to avoid the case where the grid width is too small and then the grid point is stuck in $\mathcal{Z}$ or $\mathbb{R}\setminus\mathcal{Z}$.
\paragraph{Implementation Teqniques.}
For implementation, we need to define the grid width $d(z_j)$ in a computable form.
Actually, computing $d(z_j)$ as defined in Theorem~\ref{thm:error_bound} presents a challenge since it requires the computation of the Lipschitz constant $L_i(z_j)$ of the attention score in the vicinity of $z_j$.
%
In this study, we define $d(z_j)$ as in Theorem~\ref{thm:error_bound}, by estimating the Lipschitz constant $L_i(z_j)$ using some heuristics.
Specifically, we introduce two types of heuristics based on the relative positions of the current grid point $z_j$ and $z^\mathrm{obs}$.
In the case where $z_j$ is far from $z^\mathrm{obs}$ (i.e., $|z_j-z^\mathrm{obs}|>0.1$), we assume that $f_i$ can be approximated by a linear function in the $\varepsilon_\mathrm{max}$-neighborhood of $z_j$.
Then, we conservatively set $L_i(z_j)=10|f_i^\prime(z_j)|$.
Here, we can also assume that the sign of $f_i$ does not change on the interval $[z_j,z_j+\varepsilon_\mathrm{max}]$ for $i$ such that $f_i(z_j)$ and $f_i^\prime(z_j)$ have the same sign, since $f_i$ is assumed to be approximated by a linear function.
This can be implemented by taking the $\min$ or $\max$ operation only for $i$ such that $f_i(z_j)f_i^\prime(z_j)<0$.
In contrast, in the case where $z_j$ is close to $z^\mathrm{obs}$ (i.e., $|z_j-z^\mathrm{obs}|<0.1$), $f_i$ may exhibit a flat shape or micro oscillations and tends to take values close to zero.
Note that careful consideration is required when any $f_i$ is close to zero, because it implies that the grid point $z_j$ is close to the boundary of $\mathcal{Z}$.
Therefore, it may not be reasonable to utilize the derivative of $f_i$ in the same way as above, so we assume that $L_i(z_j)=1$.
This assumption is highly conservative, since the range of the attention score $\mathcal{A}_i$ is $[0,1]$.
The schematic illustration of these heuristics is shown in Figure~\ref{fig:heuristics}.
\paragraph{Derivative of the Attention Map.}
We considered utilizing the derivative of each $f_i$ to compute the grid width $d(z_j)$.
This necessitates computing the derivative of the attention map $\mathcal{A}$, which is the output of the ViT model.
Auto differentiation, which is implemented in many deep learning frameworks (e.g., TensorFlow and PyTorch), can be used to compute this derivative.
%
%
It should be noted that we are discussing differentiating an $n$-dimensional attention map with respect to a scalar input $z_j$.
%
%
When output dimension is larger, reverse-mode auto differentiation (also called backpropagation) is generally inefficient.
%
%
In these cases, forward-mode auto differentiation is a better option.
However, it is not well supported in many frameworks and may require more implementation costs than using the back-mode auto differentiation.
%
%
We modularized the operations specific to the ViT model for differentiating the attention map using forward-mode auto differentiation in TensorFlow.
This allows us to differentiate the attention map for ViTs of any architecture without incurring additional implementation costs.
For details, see our code which is provided as a supplementary material.

\newpage
\section{Numerical Experiments}
\label{sec:sec4}
\paragraph{Methods for Comparison.}
We compared the proposed method (\texttt{adaptive}) with naive test (\texttt{naive}), permutation test (\texttt{permutation}), and bonferroni correction (\texttt{bonferroni}), in terms of type I error rate and power.
Then, we compared the proposed method with other grid search options (\texttt{fixed}, \texttt{combination}) in terms of computation time.
%
%
%
The details of the methods for comparison are described in Appendix~\ref{app:methods_for_comparison}.
\paragraph{Experimental Setup.}
%
%
%
We firstly trained the ViT classifier model on the synthetic dataset.
We made the synthetic dataset by generating 1,000 negative images $\bm{X}=(X_1,\ldots,X_n)\sim \mathcal{N}(\bm{0},I)$ and 1,000 positive images $\bm{X}=(X_1,\ldots,X_n)\sim \mathcal{N}(\bm{\mu},I)$.
The pixel intensity vector $\bm{\mu}$ is set to $\mu_i=\Delta,\ \forall i\in\mathcal{S}$ and $\mu_i=0,\ \forall i\in[n]\setminus \mathcal{S}$, where $\Delta$ is uniformly sampled from $\mathcal{U}_{[1, 4]}$ and $\mathcal{S}$ is the region to focus on whose location is randomly determined.
After training process, we experimented with the trained ViT model on the test dataset.
We input the test image to the trained ViT model and obtained the attention map, and then performed the statistical test for the obtained attention map.
In all experiments, we set the threshold value $\tau=0.6$, the grid search interval $[-S, S]$ with $S=10+|z^\mathrm{obs}|$, the minimum grid width $\varepsilon_\mathrm{min}=10^{-4}$, the maximum grid width $\varepsilon_\mathrm{max}=0.2$, and the significance level $\alpha=0.05$.
We considered the two types of covariance matrices: $\Sigma=I_n\in\mathbb{R}^{n\times n}$ (independence) and $\Sigma=(0.5^{|i-j|})_{ij}\in\mathbb{R}^{n\times n}$ (correlation).

\begin{figure}[htbp]
    \begin{minipage}[b]{0.49\linewidth}
        \centering
        \includegraphics[width=1.0\linewidth]{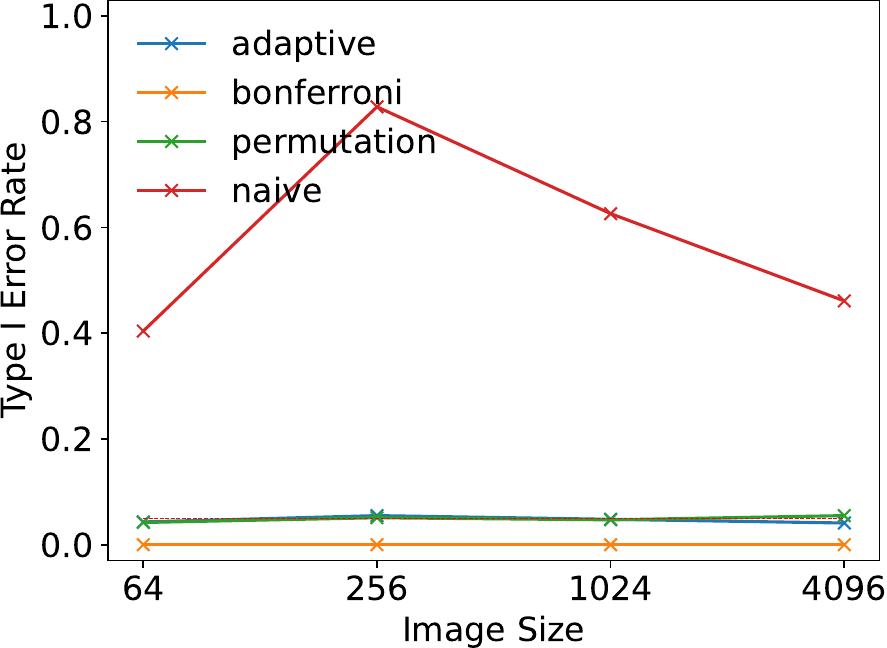}
        \subcaption{Independence}
    \end{minipage}
    \begin{minipage}[b]{0.49\linewidth}
        \centering
        \includegraphics[width=1.0\linewidth]{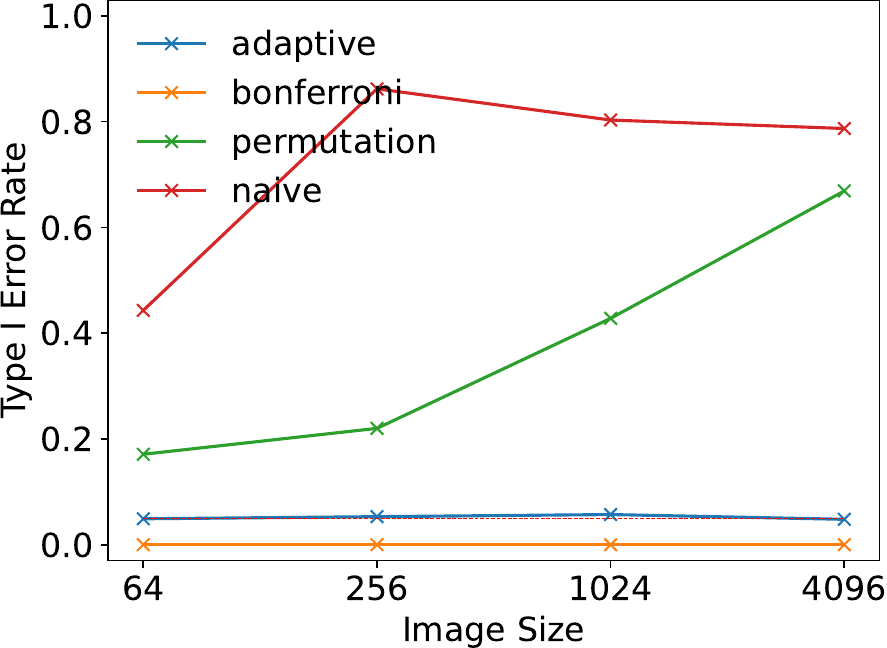}
        \subcaption{Correlation}
    \end{minipage}
    \caption{Type I Error Rate (image size)}
    \label{fig:fpr_image_size}
\end{figure}

\begin{figure}[htbp]
    \begin{minipage}[b]{0.49\linewidth}
        \centering
        \includegraphics[width=1.0\linewidth]{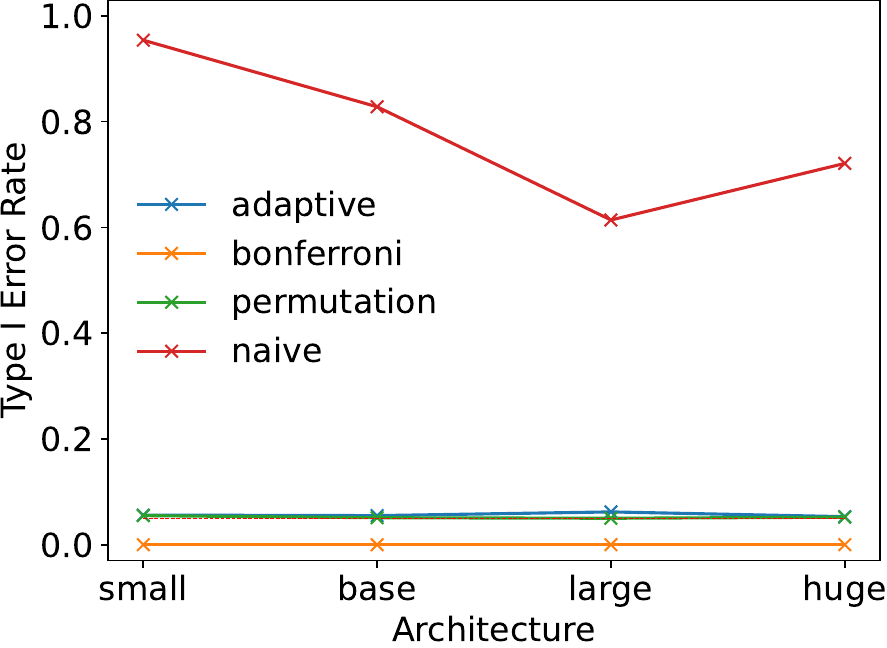}
        \subcaption{Independence}
    \end{minipage}
    \begin{minipage}[b]{0.49\linewidth}
        \centering
        \includegraphics[width=1.0\linewidth]{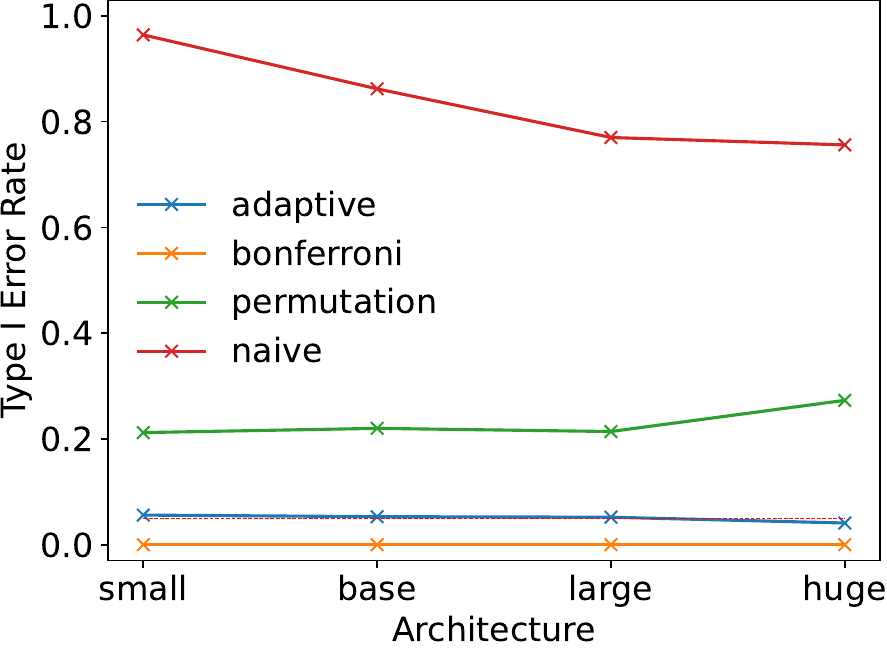}
        \subcaption{Correlation}
    \end{minipage}
    \caption{Type I Error Rate (architecture)}
    \label{fig:fpr_architecture}
\end{figure}

\begin{figure}[htbp]
    \begin{minipage}[b]{0.49\linewidth}
        \centering
        \includegraphics[width=1.0\linewidth]{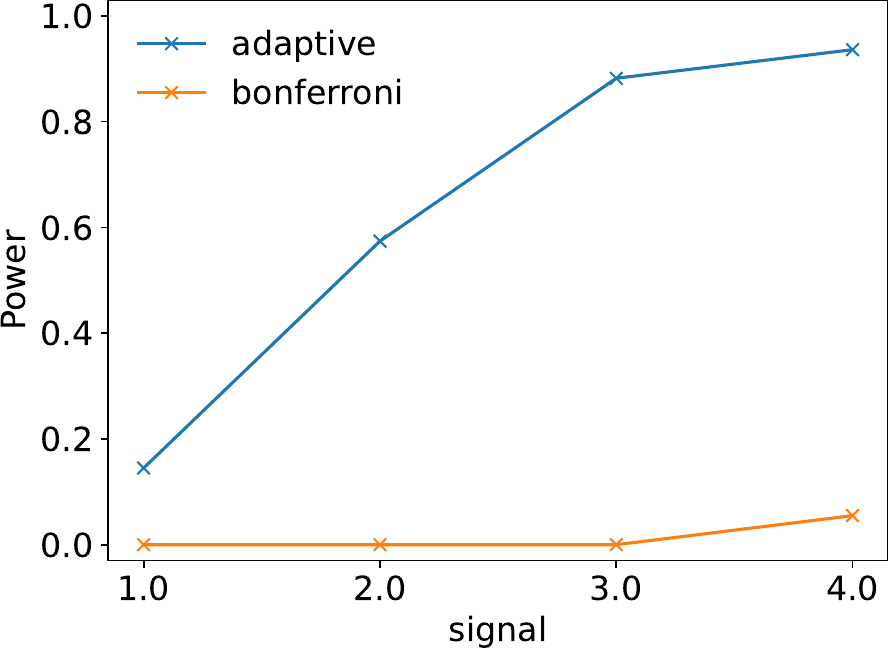}
        \subcaption{Independence}
    \end{minipage}
    \begin{minipage}[b]{0.49\linewidth}
        \centering
        \includegraphics[width=1.0\linewidth]{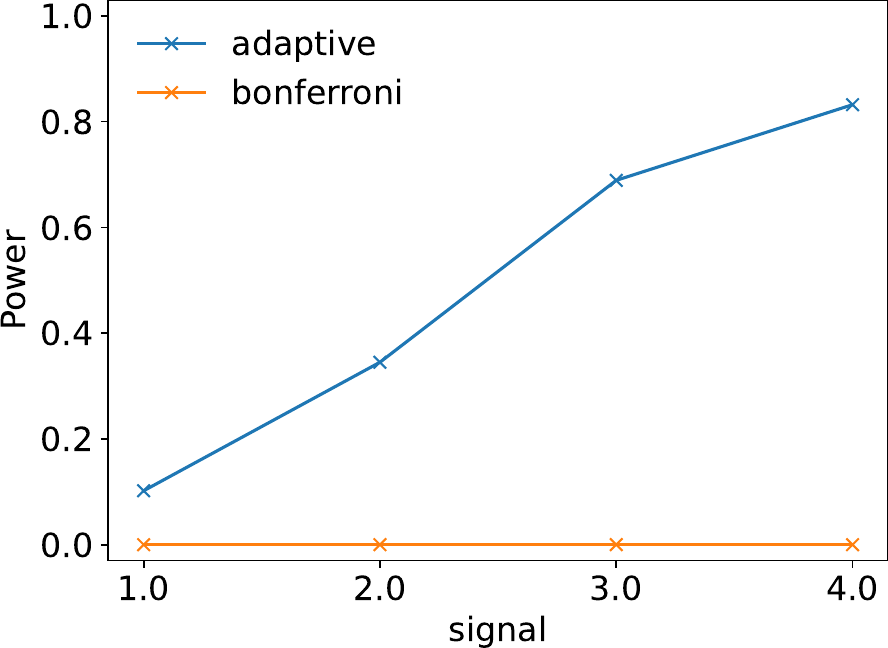}
        \subcaption{Correlation}
    \end{minipage}
    \caption{Power}
    \label{fig:alter_tpr}
\end{figure}

\begin{figure}[htbp]
    \begin{minipage}[b]{0.49\linewidth}
        \centering
        \includegraphics[width=1.0\linewidth]{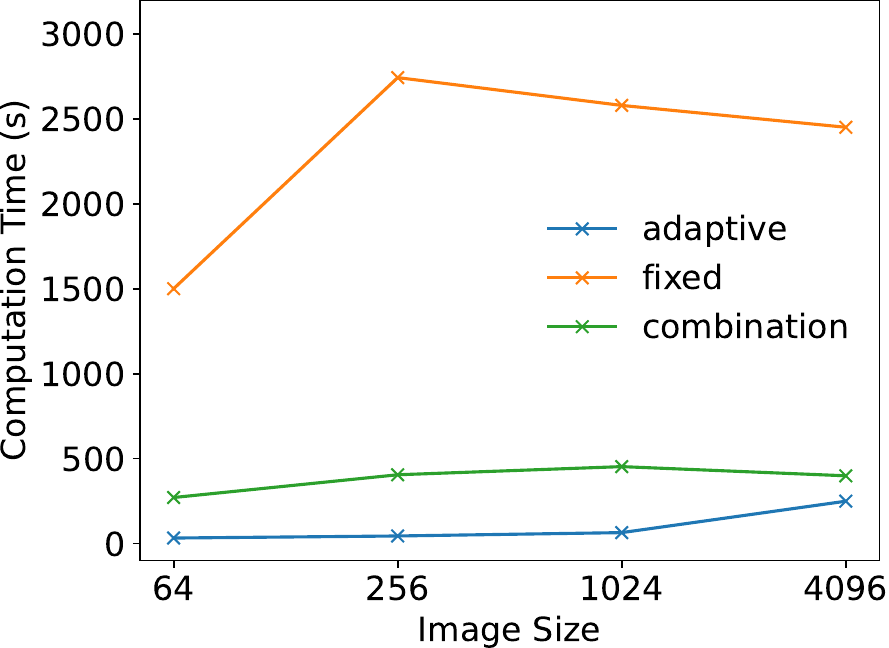}
        \subcaption{Independence}
    \end{minipage}
    \begin{minipage}[b]{0.49\linewidth}
        \centering
        \includegraphics[width=1.0\linewidth]{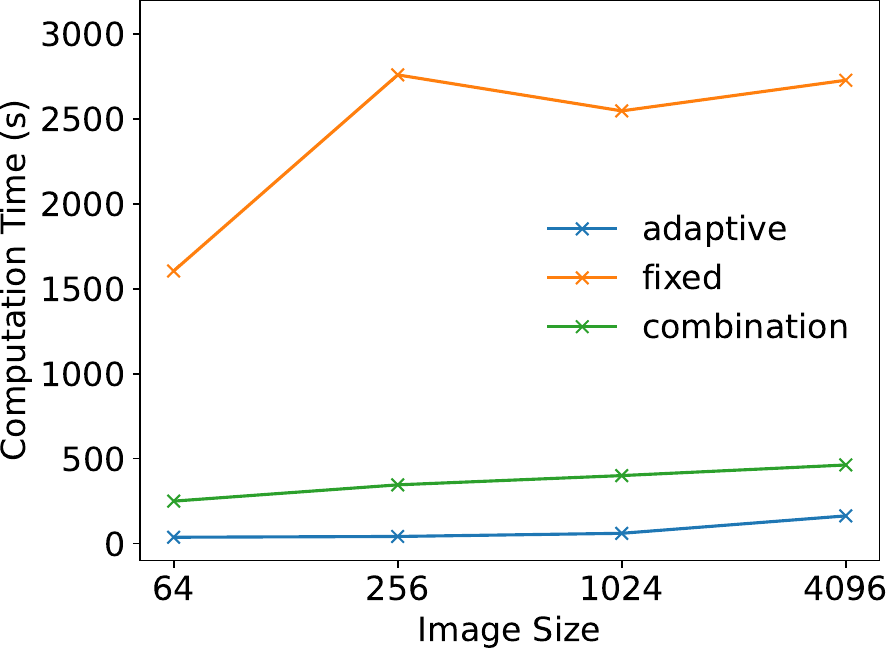}
        \subcaption{Correlation}
    \end{minipage}
    \caption{Computation Time (image size)}
    \label{fig:time_image_size}
\end{figure}

\begin{figure}[htbp]
    \begin{minipage}[b]{0.49\linewidth}
        \centering
        \includegraphics[width=1.0\linewidth]{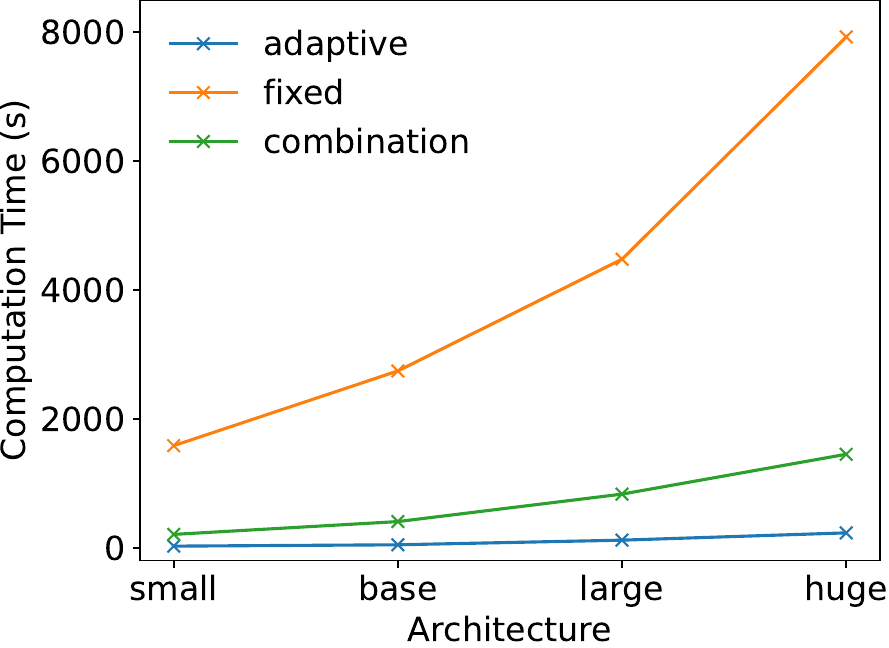}
        \subcaption{Independence}
    \end{minipage}
    \begin{minipage}[b]{0.49\linewidth}
        \centering
        \includegraphics[width=1.0\linewidth]{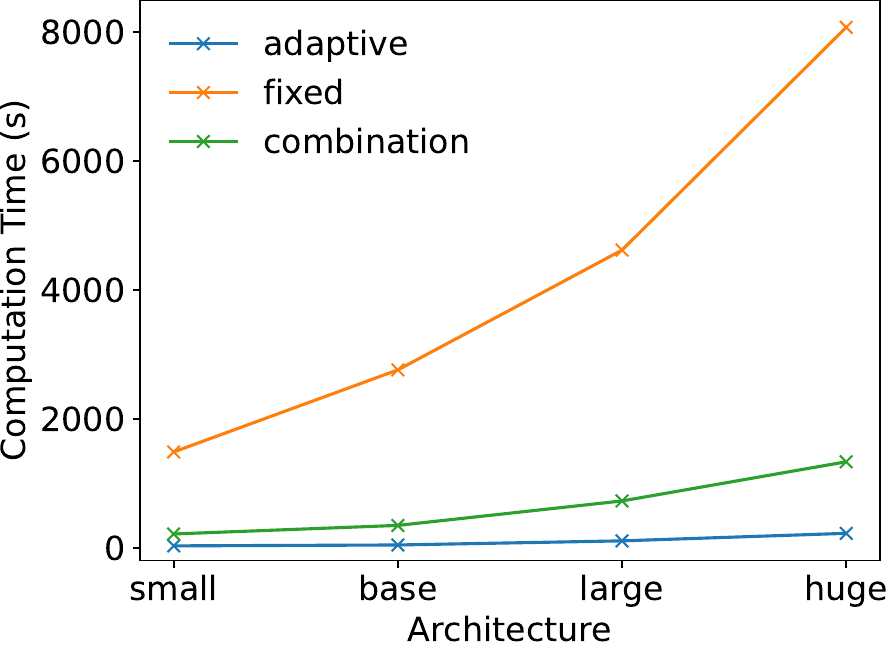}
        \subcaption{Correlation}
    \end{minipage}
    \caption{Computation Time (architecture)}
    \label{fig:time_architecture}
\end{figure}

\begin{figure}[htbp]
    \centering
    \includegraphics[width=0.7\linewidth]{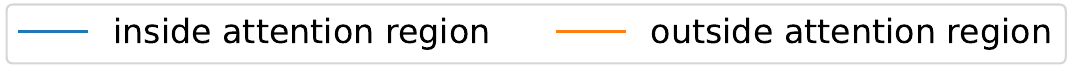}\linebreak
    \begin{minipage}[b]{0.32\linewidth}
        \centering
        \includegraphics[width=1.0\linewidth]{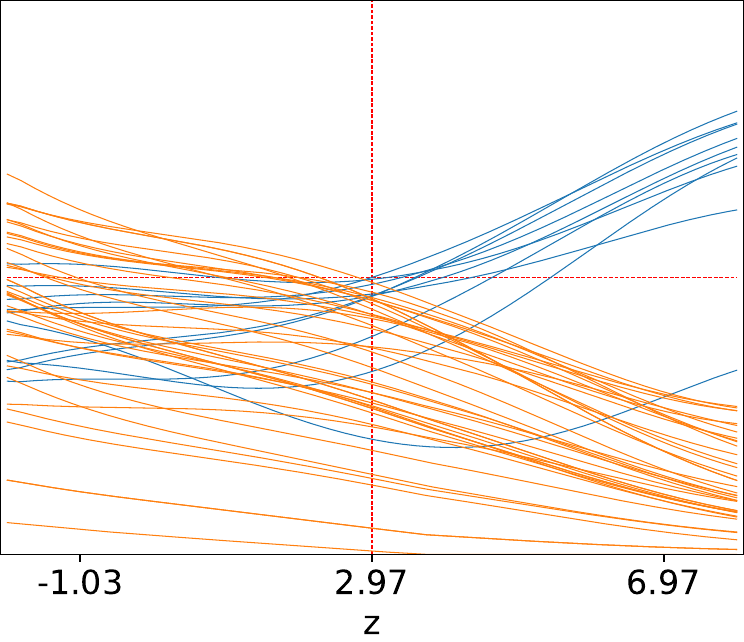}
    \end{minipage}
    \begin{minipage}[b]{0.32\linewidth}
        \centering
        \includegraphics[width=1.0\linewidth]{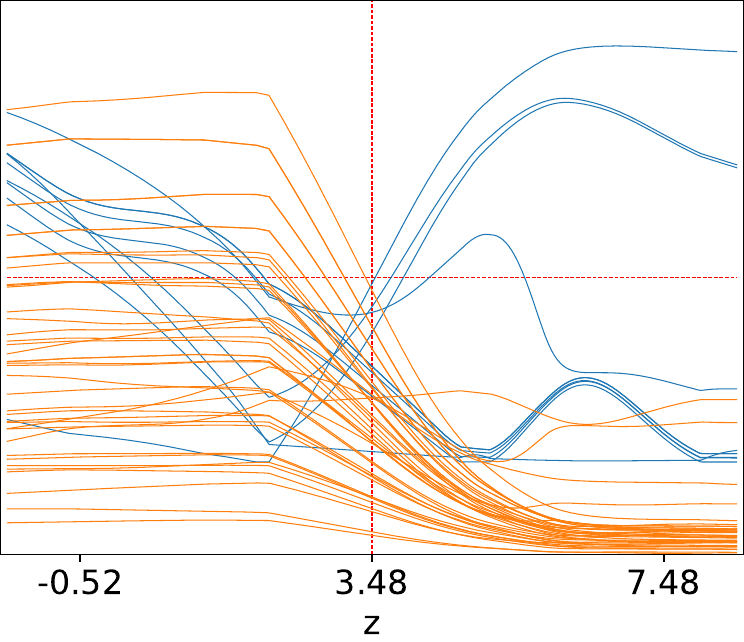}
    \end{minipage}
    \begin{minipage}[b]{0.32\linewidth}
        \centering
        \includegraphics[width=1.0\linewidth]{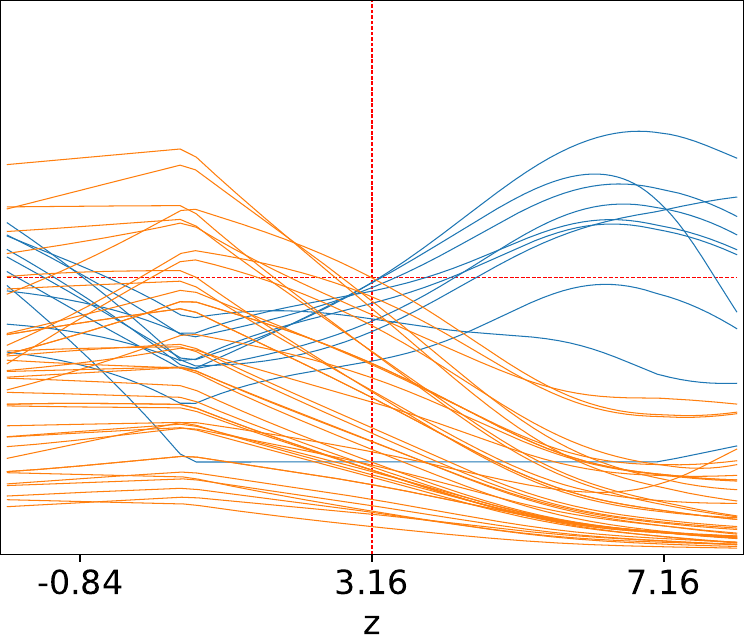}
    \end{minipage}
    \caption{
    Demonstration of the target function $f_i$.
    We set image size to 256 and architecture to base, and the image was generated from the $\mathcal{N}(\bm{0},I)$.
    The vertical red line indicates the observed test statistic $z^\mathrm{obs}$ and the horizontal red line indicates zero.
    The blue plots display $f_i$ values for 10 randomly selected $i$ in the $\mathcal{M}_{\bm{X}^\mathrm{obs}}$, while the orange plots display $f_i$ values for 40 randomly selected $i$ in the $\mathcal{M}_{\bm{X}^\mathrm{obs}}^c$.
    }
    \label{fig:demo_of_function}
\end{figure}

\begin{figure}[htbp]
    \centering
    \includegraphics[width=0.7\linewidth]{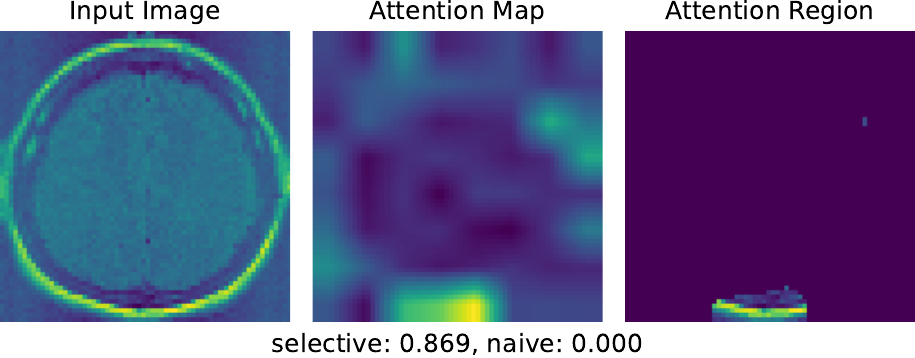}
    \includegraphics[width=0.7\linewidth]{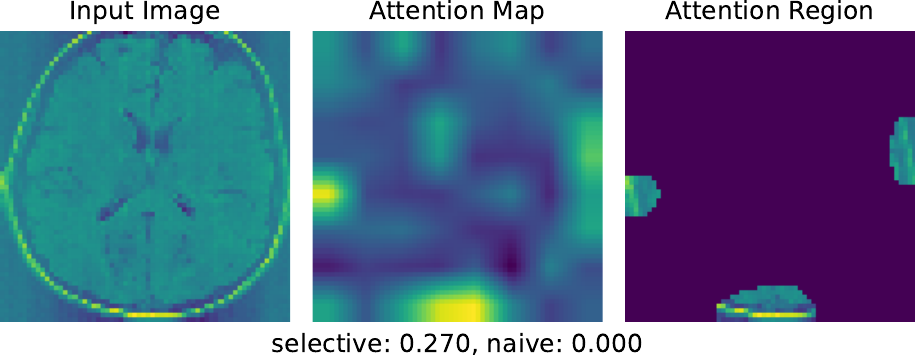}
    \includegraphics[width=0.7\linewidth]{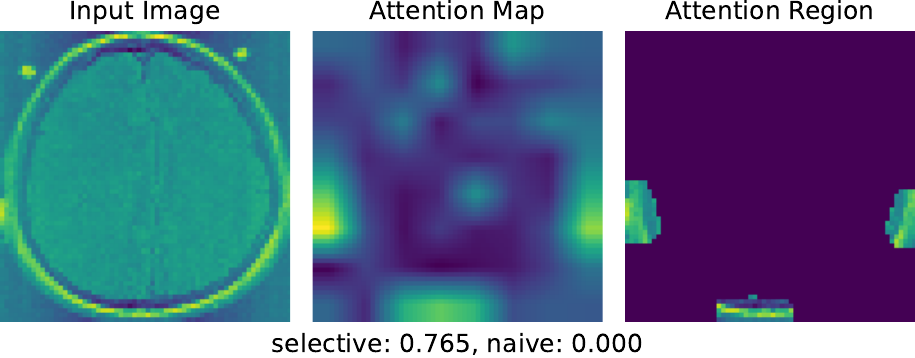}
    \includegraphics[width=0.7\linewidth]{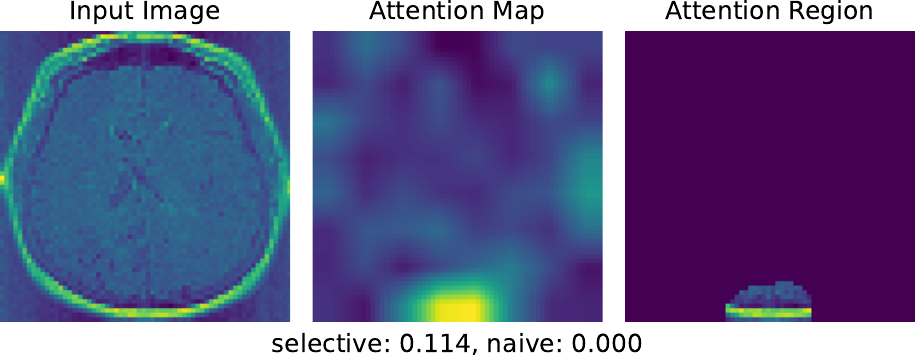}
    \includegraphics[width=0.7\linewidth]{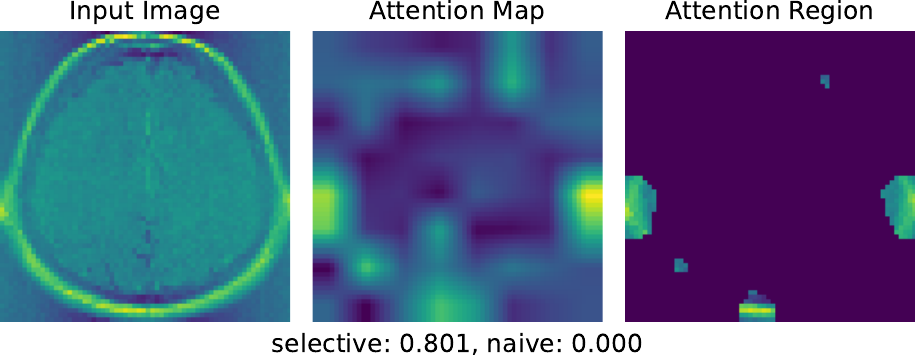}
    \caption{Demonstration on brain images without tumor}
    \label{fig:brain_without_tumor}
\end{figure}

\begin{figure}[htbp]
    \centering
    \includegraphics[width=0.7\linewidth]{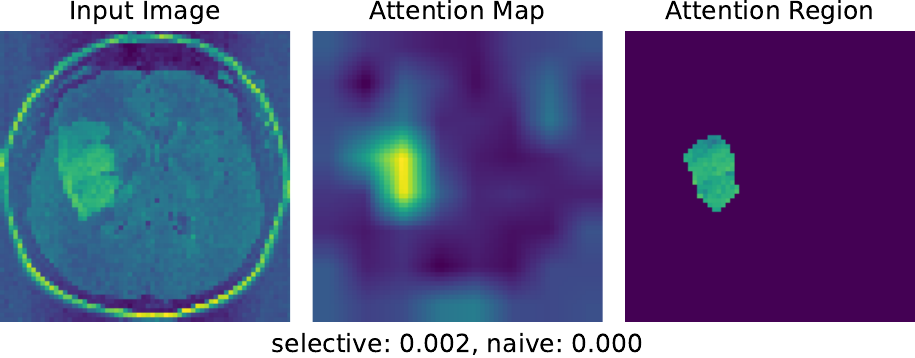}
    \includegraphics[width=0.7\linewidth]{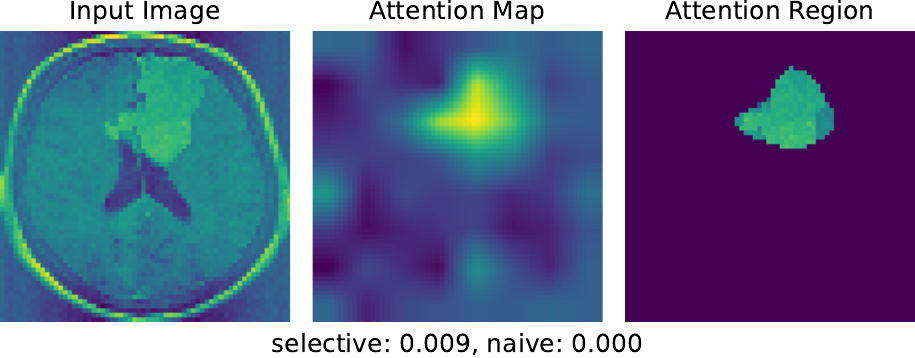}
    \includegraphics[width=0.7\linewidth]{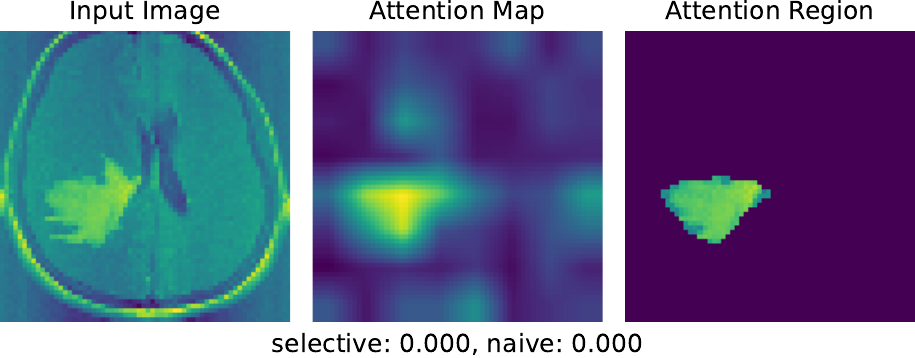}
    \includegraphics[width=0.7\linewidth]{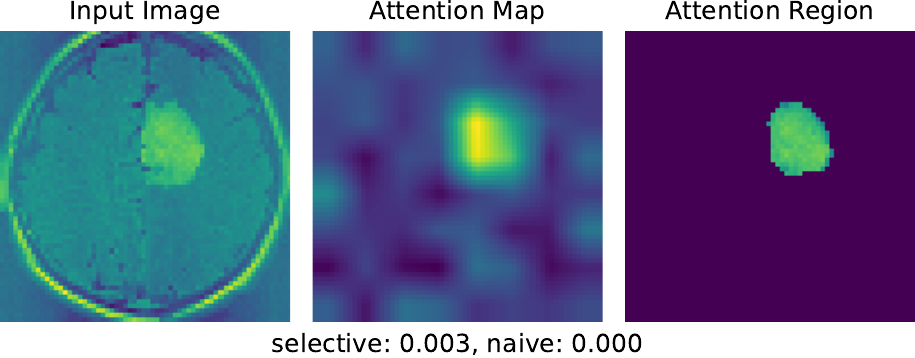}
    \includegraphics[width=0.7\linewidth]{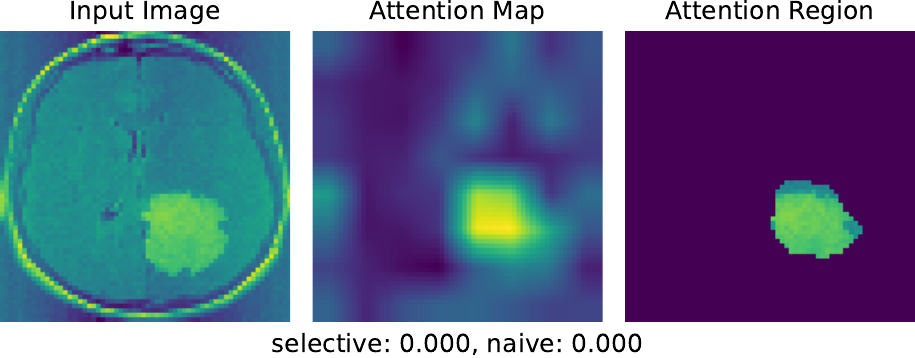}
    \caption{Demonstration on brain images with tumor}
    \label{fig:brain_with_tumor}
\end{figure}

For the experiments to see the type I error rate, we considered the two options: for image size in \{64, 256, 1024, 4096\} and for architecture in \{small, base, large, huge\} (the details of architectures in Appendix~\ref{app:architectures_to_compare}).
If not specified, we used the image size of 256 and the architecture of base.
For each setting, we generated 100 null test images $X=(X_1,\ldots,X_n)\sim \mathcal{N}(\bm{0},\Sigma)$ and run 10 trials (i.e., 1,000 null images in total).
Here, first 2 trials were also used for comparing the computation time.
Regarding our proposed method, we run additional 90 trials to check the validity in Appendix~\ref{app:additional_trials}.
%
%
%
To investigate the power, we set image size to 256 and architecture to base and generated 1,000 test images $\bm{X}=(X_1,\ldots,X_n)\sim \mathcal{N}(\bm{\mu},\Sigma)$.
The pixel intensity vector $\bm{\mu}$ is set to $\mu_i=\Delta,\ \forall i\in\mathcal{S}$ and $\mu_i=0,\ \forall i\in[n]\setminus \mathcal{S}$, where $\mathcal{S}$ is the region to focus on whose location is randomly determined.
We set $\Delta\in\{1.0,2.0,3.0,4.0\}$.
\paragraph{Results.}
The results of type I error rate are shown in Figures~\ref{fig:fpr_image_size} and \ref{fig:fpr_architecture}.
The \texttt{adaptive} and \texttt{bonferroni} successfully controlled the type I error rate under the significance level in all settings, whereas the other two methods \texttt{naive} and \texttt{permutation} could not.
Because the \texttt{naive} and \texttt{permutation} failed to control the type I error rate, we no longer considered their power.
The results of power comparison are shown in Figure~\ref{fig:alter_tpr} and we confirmed that the \texttt{adaptive} has the much higher power than the \texttt{bonferroni} in all settings.
The results of computation time for null test images are shown in Figures~\ref{fig:time_image_size} and \ref{fig:time_architecture}.
In all settings, the \texttt{adaptive} outperforms the \texttt{fixed} and \texttt{combination} while utilizing the smallest minimum grid width.
\paragraph{Discussion.}
Our experiments confirmed that the approximation approach works well with the heuristics considered in \S\ref{sec:sec3} for the attention map in the ViT model.
%
%
%
%
We assess the reasonableness of the heuristics in \S\ref{sec:sec3} by presenting several examples of our target function $f_i$ defined in~\eqref{eq:definition_of_fi} in Figure~\ref{fig:demo_of_function}.
The plots demonstrate that the function $f_i$ is generally consistent with the heuristics, having a shape that can be approximated linearly when $z$ is away from $z^\mathrm{obs}$ and tending to take values close to zero when $z$ is close to $z^\mathrm{obs}$.
The input of the ViT model is given by $\bm{a}+\bm{b}z=\bm{a}+\bm{b}z^\mathrm{obs}+\bm{b}(z-z^\mathrm{obs})=\bm{X}^\mathrm{obs}+\bm{b}(z-z^\mathrm{obs})$, where $\bm{b}$ is the vector parallel to $\bm{\eta}_{\mathcal{M}_{\bm{X}^\mathrm{obs}}}$ from the definition.
For $z$ away from $z^\mathrm{obs}$, the input $\bm{X}^\mathrm{obs}+\bm{b}(z-z^\mathrm{obs})$ results in an image where the pixel intensity in the attention region $\mathcal{M}_{\bm{X}^\mathrm{obs}}$ is highlighted.
Then, each attention score $\mathcal{A}_i$ may exhibit a gradual trend.
On the other hand, for $z$ close to $z^\mathrm{obs}$, from the definition of $f_i$ in~\eqref{eq:definition_of_fi} and continuity, it is expected that some $f_i$ are close to zero.
However, it is unclear whether these heuristics are always valid for any complex Transformer architectures.
In some cases, it may be necessary to sufficiently reduce the grid size to account for highly nonlinear functions with increased computational costs.
\paragraph{Real Data Experiments.}
We examined the brain image dataset extracted from the dataset used
in~\citep{buda2019association}, which included 939 and 941 images with and without tumors, respectively.
We selected 100 images without tumors to estimate the variance and used 700 images each with and without tumors for training the ViT classifier model.
The remaining images with and without tumors were used for testing, i.e., to demonstrate the advantages of the proposed method.
The results of the \texttt{adaptive} and \texttt{naive} are shown in Figures~\ref{fig:brain_without_tumor} and \ref{fig:brain_with_tumor}.
The naive $p$-values remain small even for images without tumors, which indicates that naive $p$-values cannot be used to quantify the reliability of the attention regions.
In contrast, the adaptive $p$-values are large for images without tumors and small for images with tumors.
This result indicates that the \texttt{adaptive} can detect true positive cases while avoiding false positive detections.

\newpage
\section{Conclusion}
\label{sec:sec5}
In this study, we introduced a novel framework for testing the statistical significance of ViT's Attention based on the concept of SI.
We provided a new computational method for computing the $p$-values as an indicators of the statistical significance.
The validity and effectiveness of the proposed method were demonstrated through its applications to synthetic data simulations and brain image diagnoses.
We believe that this study opens an important direction in ensuring the reliability of ViT's Attention.

\newpage
\subsection*{Acknowledgement}
This work was partially supported by MEXT KAKENHI (20H00601), JST CREST (JPMJCR21D3, JPMJCR22N2), JST Moonshot R\&D (JPMJMS2033-05), JST AIP Acceleration Research (JPMJCR21U2), NEDO (JPNP18002, JPNP20006) and RIKEN Center for Advanced Intelligence Project.

\clearpage
\appendix
\newpage
\section{Details of the Vision Transformer Model}
\label{app:vit}
\subsection{Structure of the Vision Transformer Model}
\label{app:vit_structure}
\begin{figure}[htbp]
    \centering
    \includegraphics[width=1.0\linewidth]{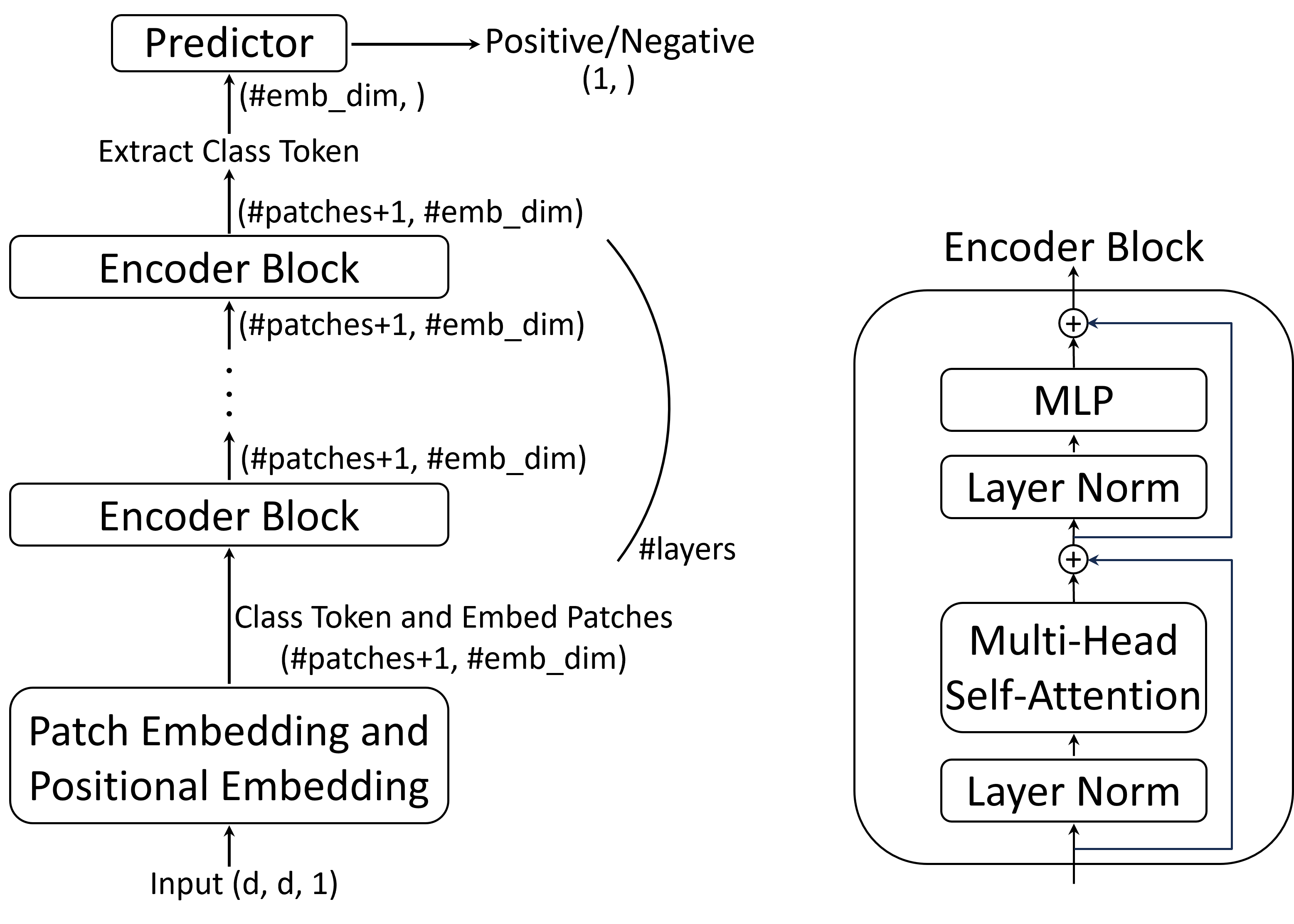}
    \caption{Structure of the Vision Transformer model.}
    \label{fig:structure}
\end{figure}

The overall structure of the ViT model is shown in Figure~\ref{fig:structure}.
In MLP, we use two fully-connected layers and set the hidden dimension to four times the \#emb\_dim.
In Multi-Head Self-Attention, we use \#heads self-attention mechanisms.
Regarding the patch embedding, we set the patch size to $\min(2,\sqrt{n}/8)$ (i.e., for $\sqrt{n}=16$ case, the patch size is $2$ and then \#patches is $(16/2)^2=64$).
As the base model, we set the \#layers to 8, the \#emb\_dim to 64, and the \#heads to 4, respectively.
\subsection{Computation of the Attention Map.}
\label{app:attention_map}
Let we denote the \#patches as $N$, \#layers as $L$, the \#heads as $H$, and the \#emb\_dim/\#heads as $D$.
%
%
We describe the computation of the attention map $\mathcal{A}(\bm{X})\in[0,1]^n$ for input image $\bm{X}\in\mathbb{R}^{n}$ from the ViT model based on~\citep{abnar2020quantifying}.
\paragraph{Obtain the Attention Weights.}
We reshape the input image $\bm{X}\in\mathbb{R}^n$ to $\bm{X}^\prime\in\mathbb{R}^{d\times d}$ where $d=\sqrt{n}$, and then input it to the ViT model.
In process of the ViT model, the input image $\bm{X}^\prime$ is passing through the self-attention mechanism $H\times L$ times.
In the $h$-th self-attention mechanism of the $l$-th layer, let we denote the query and key as $Q_{l,h}\in\mathbb{R}^{(N+1)\times D}$ and $K_{l,h}\in\mathbb{R}^{(N+1)\times D}$, respectively.
Then, the attention weights $A_{l,h}\in\mathbb{R}^{(N+1)\times (N+1)}$ are computed as
\begin{equation}
    A_{l,h}=\operatorname{softmax}\left(\frac{Q_{l,h}K_{l,h}^\top}{\sqrt{D}}\right),\
    (l, h)\in[L]\times[H],
\end{equation}
where softmax operation is applied to each row of the matrix.
Note that the row of $A_{l,h}$ corresponds to the queries and the column of $A_{l,h}$ corresponds to the keys.
The attention map is computed by aggregating the all attention weights $\{A_{l,h}\}_{(l,h)\in[L]\times[H]}$.
\paragraph{Aggregate the Attention Weights.}
We compute the layer-wise attention weights $\hat{A}_l\in\mathbb{R}^{(N+1)\times (N+1)}$ by averaging the attention weights $A_{l,h}$ in heads direction as
\begin{equation}
    \hat{A}_l=\frac{1}{H}\sum_{h\in[H]}A_{l,h},\ l\in[L].
\end{equation}
Then, to aggregate the all attention weights to $\bar{A}$, we take the matrix product of each $\hat{A}_l$, adding the identity matrix $I\in\mathbb{R}^{(N+1)\times (N+1)}$, as
\begin{equation}
    \bar{A}=\prod_{l\in[L]}\left(\hat{A}_l+I\right),
\end{equation}
where matrix $I$ represents the skip connection in Encoder Block as in Figure~\ref{fig:structure}.
Finally, we extract the $N$-dimensional vector $A\in\mathbb{R}^N$ from $\bar{A}$ as
\begin{equation}
    A=\bar{A}_{1,2:N+1},
\end{equation}
which corresponds to the keys of each patch for the query of the class token as an aggregated form.
\paragraph{Post-Processing.}
We reshape the $N$-dimensional vector $A$ to square matrix and upscale it to $A^\prime$ whose size is the same as the input image $\bm{X}^\prime$ by using bilinear interpolation.
Then, we obtain the attention map $\mathcal{A}(\bm{X})\in[0,1]^n$ by normalizing $A^\prime$ with min-max normalization and flattening it to $n$-dimensional vector.
\newpage
\section{Proofs}
\label{app:proof}
\subsection{Proof of Theorem \ref{thm:error_bound}}
\label{app:proof_of_thm}
We note that the $\varepsilon_\mathrm{max}$ is not necessarily to evaluate the error bound because it is introduced for implementation convenience.
Let us define the indicator function $I(z_j)$ as
\begin{equation}
    I(z_j) =
    \begin{cases}
        1 & (\varepsilon_\mathrm{min}\leq d(z_j)) \\
        0 & (\varepsilon_\mathrm{min}> d(z_j))
    \end{cases}
\end{equation}
First, we divide $\mathbb{R}$ into the four unions of intervals such that any two of them have no intersection with length as
\begin{align}
    R^1 & =
    \bigcup_{j\mid I(z_j)=1, z_j\in\mathcal{Z}}
    [z_j, z_{j+1}]  \cup J(z^\mathrm{obs}),                                      \\
    R^2 & = \bigcup_{j\mid I(z_j)=1, z_j\notin\mathcal{Z}} [z_j, z_{j+1}],       \\
    R^3 & = \bigcup_{j\mid I(z_j)=0} [z_j, z_{j+1}] \setminus J(z^\mathrm{obs}), \\
    R^4 & = (-\infty, -S]\cup [S, \infty).
\end{align}
Here, $R^1\subset \mathcal{Z}^\mathrm{grid}$ and $R^2\subset \mathbb{R}\setminus\mathcal{Z}^\mathrm{grid}$ are obvious from the definition of them, and from the Lemma~\ref{lem:grid_width}, we have $R^1\subset \mathcal{Z}$ and $R^2\subset \mathbb{R}\setminus\mathcal{Z}$.
Then, we have following subset relationships
\begin{equation}
    \label{eq:thm:subset_relationship}
    R^1\subset \mathcal{Z},\ \mathcal{Z}^\mathrm{grid} \subset R^1\cup R^3\cup R^4
\end{equation}

Let us denote the probability density function of the standard Gaussian distribution as $\phi$ and the cumulative distribution function of that as $\Phi$, and introduce the integrate function $\mathcal{I}$ as
\begin{equation}
    \mathcal{I}:\mathcal{B}(\mathbb{R})\ni R\mapsto \int_{R}\phi(z)dz\in[0, 1],
\end{equation}
where $\mathcal{B}(\mathbb{R})$ is the Borel set of $\mathbb{R}$.
Additionally, for any $R\in\mathcal{B}(\mathbb{R})$, we define the two sets $R_\mathrm{in}, R_\mathrm{out}\in \mathcal{B}(\mathbb{R})$ as
\begin{equation}
    R_\mathrm{in} = R\cap [-|z^\mathrm{obs}|, |z^\mathrm{obs}|],\
    R_\mathrm{out} = R\setminus[-|z^\mathrm{obs}|, |z^\mathrm{obs}|].
\end{equation}
Then, we have $p_\mathrm{selective}$ and $p_\mathrm{grid}$ as
\begin{gather}
    \label{eq:thm:p_selective}
    p_\mathrm{selective}
    =
    \mathbb{P}_{\mathrm{H}_0}
    (|Z|>|z^\mathrm{obs}|\mid Z\in\mathcal{Z})
    =
    \frac
    {\mathcal{I}(\mathcal{Z}_\mathrm{out})}
    {\mathcal{I}(\mathcal{Z}_\mathrm{in})},\\
    \label{eq:thm:p_grid}
    p_\mathrm{grid}
    =
    \mathbb{P}_{\mathrm{H}_0}
    (|Z|>|z^\mathrm{obs}|\mid Z\in\mathcal{Z}^\mathrm{grid})
    =
    \frac
    {\mathcal{I}(\mathcal{Z}^\mathrm{grid}_\mathrm{out})}
    {\mathcal{I}(\mathcal{Z}^\mathrm{grid}_\mathrm{in})},
\end{gather}
respectively.
Therefore, by considering the subset relationships in~\eqref{eq:thm:subset_relationship}, our goal of evaluating the error is casted into the evaluating the $\mathcal{I}(R^1)$, $\mathcal{I}(R^3)$, and $\mathcal{I}(R^4)$.
To do so, we start to evaluate the length of $R^3$.

We denote the Lipschitz constant of $f_i$ as $L_i>0$ and the number of zeros of $f_i$ as $K_i\in\mathbb{N}$.
We define the $L>0$ and $K\in\mathbb{N}$ as $L = \max_{i\in[n]}L_i$ and $K = \max_{i\in[n]}K_i$, respectively.
Then, for any $z_j\in\mathcal{Z}$, we have
\begin{equation}
    d(z_j)\geq \min_{i\in[n]}\frac{|f_i(z_j)|}{L_i(z_j)}\geq
    \min_{i\in[n]}\frac{|f_i(z_j)|}{L}
\end{equation}
Furthermore, regarding the condition of $R^3$, we have $\varepsilon_\mathrm{min}> \min_{i\in[n]}|f_i(z_j)|/L$ from $I(z_j)=0\Leftrightarrow \varepsilon_\mathrm{min}>d(z_j)$.
Therefore, we have the following subset relationship
\begin{align}
    R^3
     & \subset
    \bigcup_{j\mid I(z_j)=0} [z_j, z_{j+1}]
    =
    \bigcup_{j\mid I(z_j)=0} [z_j, z_j+\varepsilon_\mathrm{min}] \\
    \label{eq:thm:R3}
     & \subset
    \bigcup_{j\mid L\varepsilon_\mathrm{min}>\min_{i\in[n]}|f_i(z_j)|}
    [z_j, z_j+\varepsilon_\mathrm{min}].
\end{align}

Continuously, we evaluate the length of $R^3$ by show that the set in~\eqref{eq:thm:R3} is restricted to the neighborhood of the zeros of $f_i$.
For $i\in[n]$, we denote the $k$-th zeros of $f_i$ as $q_{ik} (k\in[K_i])$, and the minimum value of $|f_i^\prime|$ at the zeros of $f_i$ as $h_i>0$ (i.e., $h_i=\min_{k\in[K_i]}|f_i^\prime(q_{ik})|$).
Let us denote the $h>0$ as $h=\min_{i\in[n]}h_i$.
Here, by using these zeros, we define the set $D(r)$ for any $r>0$, which is the union of the $r$-neighborhood of the zeros,
\begin{equation}
    D(r) = \bigcup_{i\in[n]}\bigcup_{k\in[K_i]}
    [q_{ik}-r, q_{ik}+r].
\end{equation}
Then, for any $i\in[n]$ and $k\in[K_i]$, from the definition of derivative function, there exists $\delta_{ik}>0$ such that, for any $s$ satisfying $0<|s|<\delta_{ik}$,
\begin{equation}
    \left|
    \frac{f_i(q_{ik}+s)-f_i(q_{ik})}{s}  - f_i^\prime(q_{ik})
    \right|
    < \frac{h}{2}
\end{equation}
holds.
Therefore, from the triangle inequality and the definition of $h$, we have
\begin{equation}
    \frac{h}{2}
    >
    \left|
    f^\prime(q_{ik}) - \frac{f_i(q_{ik}+s)}{s}
    \right|
    \geq
    \left|
    f^\prime(q_{ik})
    \right|
    -
    \left|
    \frac{f_i(q_{ik}+s)}{s}
    \right|
    \geq
    h -
    \left|
    \frac{f_i(q_{ik}+s)}{s}
    \right|.
\end{equation}
To summarize, we have $|f_i(q_{ik}+s)|\geq h|s|/2$ including the case of $s=0$.
Thus, let us denote the $\delta>0$ as $\delta = \min_{i\in[n]}\min_{k\in[K_i]}\delta_{ik}$, then, for any $s$ satisfying $|s|<\delta$, we have
\begin{equation}
    \label{eq:thm:lower_of_f}
    \min_{i\in[n]}\min_{k\in[K_i]}|f_i(q_{ik}+s)|\geq \frac{h}{2}|s|.
\end{equation}
Next, we consider the set $[-S,S]\setminus D(\delta)$, which is assumed to have its boundary points added.
Then, this set is a compact set, and thus the minimum value of $\min_{i\in[n]}|f_i|$ in this set is attained and we denote it as $l>0$ (because $l=0$ violates the assumption of zeros of $f_i$ and the definition of $D(\delta)$).

As follows, we consider the asymptotic case of $\varepsilon_\mathrm{min}\to 0$ and then only consider the case of $\varepsilon_\mathrm{min}<\min(h\delta/2L, l/L)$.
In this case, we have $0<2L\varepsilon_\mathrm{min}/h<\delta$, thus, from~\eqref{eq:thm:lower_of_f}, the infimum of $\min_{i\in[n]}|f_i|$ in $D(\delta)/D(2L\varepsilon_\mathrm{min}/h)$ is greater than or equal to $h(2L\varepsilon_\mathrm{min}/h)/2=L\varepsilon_\mathrm{min}$.
By combining this with the definition of $l$, for any $z\in[-S,S]\setminus D(2L\varepsilon_\mathrm{min}/h)$, we have
\begin{equation}
    \min_{i\in[n]}|f_i(z)|\geq
    \min(L\varepsilon_\mathrm{min},l)
    =
    L\varepsilon_\mathrm{min},
\end{equation}
where we used the assumption of $\varepsilon_\mathrm{min}<l/L$.
Therefore, we have
\begin{align}
    R^3
     & \subset
    \bigcup_{j\mid L\varepsilon_\mathrm{min}>\min_{i\in[n]}|f_i(z_j)|}
    [z_j, z_j+\varepsilon_\mathrm{min}] \\
     & \subset
    \bigcup_{j\mid z_j\in D(2L\varepsilon_\mathrm{min}/h)}
    [z_j, z_j+\varepsilon_\mathrm{min}]
    \subset
    D\left(\left(\frac{2L}{h}+1\right)\varepsilon_\mathrm{min}\right).
\end{align}

Based on these results, we return to the evaluation of the $\mathcal{I}(R^1)$, $\mathcal{I}(R^3)$, and $\mathcal{I}(R^4)$.
Regarding the $\mathcal{I}(R^3)$, we have
\begin{align}
    \mathcal{I}(R^3)
     & \leq
    \mathcal{I}
    \left(D\left(\left(\frac{2L}{h}+1\right)\varepsilon_\mathrm{min}\right)\right) \\
     & \leq
    \sum_{i\in[n]}\sum_{k\in[K_i]}
    \mathcal{I}\left(
    \left[
        q_{ik}-\left(\left(\frac{2L}{h}+1\right)\varepsilon_\mathrm{min}\right),
        q_{ik}+\left(\left(\frac{2L}{h}+1\right)\varepsilon_\mathrm{min}\right)
        \right]
    \right)                                                                        \\
     & = \sum_{i\in[n]}\sum_{k\in[K_i]}
    \left\{
    \Phi\left(q_{ik}+\left(\frac{2L}{h}+1\right)\varepsilon_\mathrm{min}\right)
    -
    \Phi\left(q_{ik}-\left(\frac{2L}{h}+1\right)\varepsilon_\mathrm{min}\right)
    \right\}.
\end{align}
By using the mean value theorem and the fact that $\phi$ has the maximum value at $0$, then we have
\begin{align}
    \mathcal{I}(R^3)
     & \leq \sum_{i\in[n]}\sum_{k\in[K_i]}
    \phi(0)\left(\frac{4L}{h}+2\right)\varepsilon_\mathrm{min}                                          \\
    \label{eq:thm:R3_bound}
     & \leq nK\phi(0)\left(\frac{4L}{h}+2\right)\varepsilon_\mathrm{min} = M_1\varepsilon_\mathrm{min},
\end{align}
where $M_1 = nK\phi(0)\left({4L}/{h}+2\right)$ is a positive constant independent of $\varepsilon_\mathrm{min}$ and $S$.
Next, regarding the $\mathcal{I}(R^1)$, from the mean value theorem, the symmetry of $\phi$ and the decreasing property of $\phi$ on $[0, \infty)$, we have
\begin{align}
    \mathcal{I}(R^1_\mathrm{in})
     & \geq
    \mathcal{I}(J(z^\mathrm{obs})_\mathrm{in})                                       \\
     & =  \mathcal{I}([z^\mathrm{obs}-d^\mathrm{obs}, z^\mathrm{obs}+d^\mathrm{obs}]
    \cap [-|z^\mathrm{obs}|, |z^\mathrm{obs}|])                                      \\
    \label{eq:thm:R1_bound}
     & \geq \phi(z^\mathrm{obs})d^\mathrm{obs} = M_2,
\end{align}
where $M_2 = \phi(z^\mathrm{obs})d^\mathrm{obs}$ is a positive constant independent of $\varepsilon_\mathrm{min}$ and $S$.
Finally, regarding the $\mathcal{I}(R^4)$, we have
\begin{equation}
    \label{eq:thm:R4_bound}
    \mathcal{I}(R^4) = 2\Phi(-S)
\end{equation}

Finally, we evaluate the error bound.
From~\eqref{eq:thm:subset_relationship}, \eqref{eq:thm:p_selective} and~\eqref{eq:thm:p_grid}, we have
\begin{equation}
    \frac{\mathcal{I}(R^1_\mathrm{out})}{\mathcal{I}((R^1\cup R^3 \cup R^4)_\mathrm{in})}
    \leq
    p_\mathrm{selective},\ p_\mathrm{grid}
    \leq
    \frac{\mathcal{I}((R^1\cup R^3 \cup R^4)_\mathrm{out})}{\mathcal{I}(R^1_\mathrm{in})}.
\end{equation}
Therefore, by using~\eqref{eq:thm:R3_bound}, \eqref{eq:thm:R1_bound} and \eqref{eq:thm:R4_bound}, we have the following error bound
\begin{align}
         & |p_\mathrm{selective}-p_\mathrm{grid}|                                         \\
    \leq &
    \frac{\mathcal{I}((R^1\cup R^3 \cup R^4)_\mathrm{out})}{\mathcal{I}(R^1_\mathrm{in})}
    -
    \frac{\mathcal{I}(R^1_\mathrm{out})}{\mathcal{I}((R^1\cup R^3 \cup R^4)_\mathrm{in})} \\
    =    &
    \frac{
    \mathcal{I}((R^1\cup R^3 \cup R^4)_\mathrm{out})\mathcal{I}((R^1\cup R^3 \cup R^4)_\mathrm{in})
    -
    \mathcal{I}(R^1_\mathrm{out})\mathcal{I}(R^1_\mathrm{in})
    }
    {\mathcal{I}(R^1_\mathrm{in})\mathcal{I}((R^1\cup R^3 \cup R^4)_\mathrm{in})}         \\
    =    &
    \frac{
    \mathcal{I}((R^3 \cup R^4)_\mathrm{out})\mathcal{I}((R^3 \cup R^4)_\mathrm{in})
    +
    \mathcal{I}(R^1_\mathrm{out})\mathcal{I}(R^3 \cup R^4)
    }
    {\mathcal{I}(R^1_\mathrm{in})\mathcal{I}((R^1\cup R^3 \cup R^4)_\mathrm{in})}         \\
    \leq &
    \frac{\mathcal{I}(R^3 \cup R^4)^2 + \mathcal{I}(R^3 \cup R^4)}
    {\mathcal{I}(R^1_\mathrm{in})^2}                                                      \\
    \label{eq:thm:error_bound}
    \leq &
    \frac{2}{M_2^2}\mathcal{I}(R^3 \cup R^4)
    \leq
    \frac{2}{M_2^2}(M_1\varepsilon_\mathrm{min} + 2\Phi(-S)).
\end{align}
Here, based on the three equations $\lim_{x\to \infty}\phi(x)/x\phi(x)=0$, $\Phi^\prime(-x)=(1-\Phi(x))^\prime=-\phi(x)$ and $\phi^\prime(x)=-x\phi(x)$, we have the $\lim_{x\to\infty}\Phi(-x)/\phi(x) = 0$ from the l'H\^opital's rule.
We consider the asymptotic case of $S\to\infty$ and then only consider the case of $S$ sufficiently large such that $\Phi(-S)\leq \exp(-S^2/2)$ holds (we can take such $S$ because $\lim_{x\to\infty}\Phi(-x)/\phi(x) = 0$).
By combining this with~\eqref{eq:thm:error_bound}, we have
\begin{align}
    |p_\mathrm{selective}-p_\mathrm{grid}|
     & \leq
    \frac{2}{M_2^2}(M_1\varepsilon_\mathrm{min} + 2\exp(-S^2/2))           \\
     & \leq \frac{2M_1+4}{M_2^2}(\varepsilon_\mathrm{min} + \exp(-S^2/2)),
\end{align}
where the coefficient $(2M_1+4)/M_2^2$ is positive constant independent of $\varepsilon_\mathrm{min}$ and $S$.
Thus, we have successfully showed that the error is bounded by $O(\varepsilon_\mathrm{min}+\exp(-S^2/2))$ in asymptotic case of $\varepsilon_\mathrm{min}\to 0$ and $S\to\infty$, which was what we wanted.
\subsection{Proof of Lemma \ref{lem:grid_width}}
\label{app:proof_of_lem}
In case of $z_j\in\mathcal{Z}$, we have
\begin{align}
    [z_j, z_j+\min(\varepsilon_\mathrm{max}, d(z_j))]
     & =
    [z_j, z_j+\varepsilon_\mathrm{max}] \cap
    \left[z_j, z_j+\min_{i\in[n],f_i(z_j)< 0}\frac{|f_i(z_j)|}{L_i(z_j)}\right] \\
     & =
    \bigcap_{i\in[n],f_i(z_j)< 0}
    [z_j, z_j+\varepsilon_\mathrm{max}]
    \cap
    \left[z_j, z_j+\frac{|f_i(z_j)|}{L_i(z_j)}\right]                           \\
     & =
    \bigcap_{i\in[n],f_i(z_j)< 0}
    \left[
        z_j, z_j+\min\left(\varepsilon_\mathrm{max}, \frac{|f_i(z_j)|}{L_i(z_j)}\right)
    \right]                                                                     \\
     & \subset \mathcal{Z}.
\end{align}
Similarly, in case of $z_j\notin\mathcal{Z}$, we have
\begin{align}
    [z_j, z_j+\min(\varepsilon_\mathrm{max}, d(z_j))]
     & =
    [z_j, z_j+\varepsilon_\mathrm{max}] \cap
    \left[z_j, z_j+\max_{i\in[n],f_i(z_j)\geq 0}\frac{|f_i(z_j)|}{L_i(z_j)}\right] \\
     & =
    \bigcup_{i\in[n],f_i(z_j)\geq 0}
    [z_j, z_j+\varepsilon_\mathrm{max}]
    \cap
    \left[z_j, z_j+\frac{|f_i(z_j)|}{L_i(z_j)}\right]                              \\
     & =
    \bigcup_{i\in[n],f_i(z_j)\geq 0}
    \left[
        z_j, z_j+\min\left(\varepsilon_\mathrm{max}, \frac{|f_i(z_j)|}{L_i(z_j)}\right)
    \right]                                                                        \\
     & \subset \mathbb{R}\setminus\mathcal{Z}.
\end{align}
\newpage
\section{Experimental Details}
\label{app:experiment}
\subsection{Methods for Comparison}
\label{app:methods_for_comparison}
We compared our proposed method with the following methods:
\begin{itemize}
    \item
          \texttt{naive}: This method uses a classical $z$-test without conditioning, i.e., we compute the naive $p$-value as
          \begin{equation*}
              p_\mathrm{naive} = \mathbb{P}_{\mathrm{H}_{0}}
              \left(
              |Z| > |z^\mathrm{obs}|
              \right).
          \end{equation*}
    \item
          \texttt{permutation}: This method uses a permutation test. The procedure is as follows:
          \begin{itemize}
              \item Compute the observed test statistic $z^\mathrm{obs}$ by inputting the observed image $\bm{X}^\mathrm{obs}$ to the ViT model.
              \item For $i=1,\ldots,B$, compute the test statistic $z^{(i)}$ by inputting the permuted image $\bm{X}^{(i)}$ to the ViT model. Here, $B$ is the number of permutations which is set to 1,000 in our experiments.
              \item Compute the permutation $p$-value as
                    \begin{equation*}
                        p_\mathrm{permutation}
                        =
                        \frac{1}{B}\sum_{b\in[B]}\bm{1}
                        \{|z^{(b)}| > |z^\mathrm{obs}|\},
                    \end{equation*}
                    where $\bm{1}\{\cdot\}$ is the indicator function.
          \end{itemize}
    \item
          \texttt{bonferroni}: This is a method to control the type I error rate by using the Bonferroni correction. The number of all possible attention regions is $2^n$, then we compute the bonferroni $p$-value as $p_\mathrm{bonferroni}=\min(1, 2^n\cdot p_\mathrm{naive})$.
    \item
          \texttt{fixed}: This is a grid search method with a fixed grid width $\varepsilon=10^{-3}$.
    \item
          \texttt{combination}: This is a grid search method with a fixed grid width $\varepsilon=10^{-4}$ for the grid point $z_j$ which satisfies $|z_j-z^\mathrm{obs}|<0.1$ and $\varepsilon=10^{-2}$ for the remaining grid points.
\end{itemize}
We note that, in implementing the grid search methods (\texttt{adaptive}, \texttt{fixed}, \texttt{combination}),
the binary search is performed to find the boundary of the truncated region $\mathcal{Z}$ between adjacent grid points, where one belongs to the $\mathcal{Z}$ and the other does not.
\subsection{Architectures to Compare}
\label{app:architectures_to_compare}
The architectures to compare are shown in Table~\ref{tab:architectures}.
The details of the structure of the ViT model we used are shown in Appendix~\ref{app:vit_structure}.
\begin{table}[htbp]
    \centering
    \caption{Architectures to compare}
    \label{tab:architectures}
    \begin{tabular}{ccccc}
        \hline
        Model & \#layers & \#hidden\_dim & \#heads & \#parameters \\ \hline
        small & 4        & 32            & 2       & 53.2K        \\
        base  & 8        & 64            & 4       & 405K         \\
        large & 12       & 128           & 8       & 2.39M        \\
        huge  & 16       & 256           & 16      & 12.7M        \\ \hline
    \end{tabular}
\end{table}
\newpage
\section{Robustness of Type I Error Rate Control}
\label{app:robustness}
In this experiment, we confirmed the robustness of the proposed method in terms of type I error rate control by applying our method to the two cases: the case where the variance is estimated from the same data and the case where the noise is non-Gaussian.
\subsection{Estimated Variance.}
\begin{figure}[htbp]
    \begin{minipage}[b]{0.49\linewidth}
        \centering
        \includegraphics[width=1.0\linewidth]{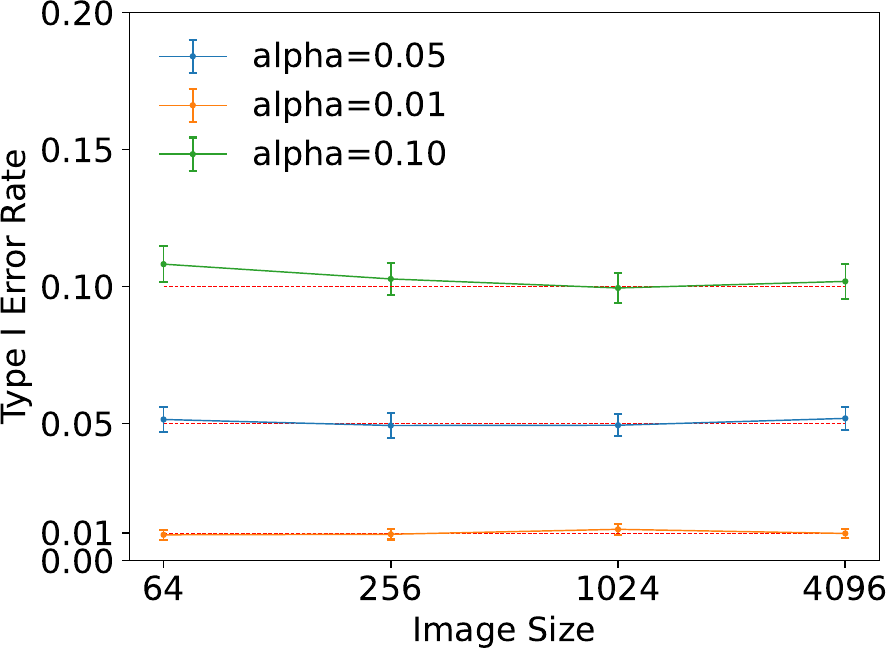}
        \subcaption{Image Size}
    \end{minipage}
    \begin{minipage}[b]{0.49\linewidth}
        \centering
        \includegraphics[width=1.0\linewidth]{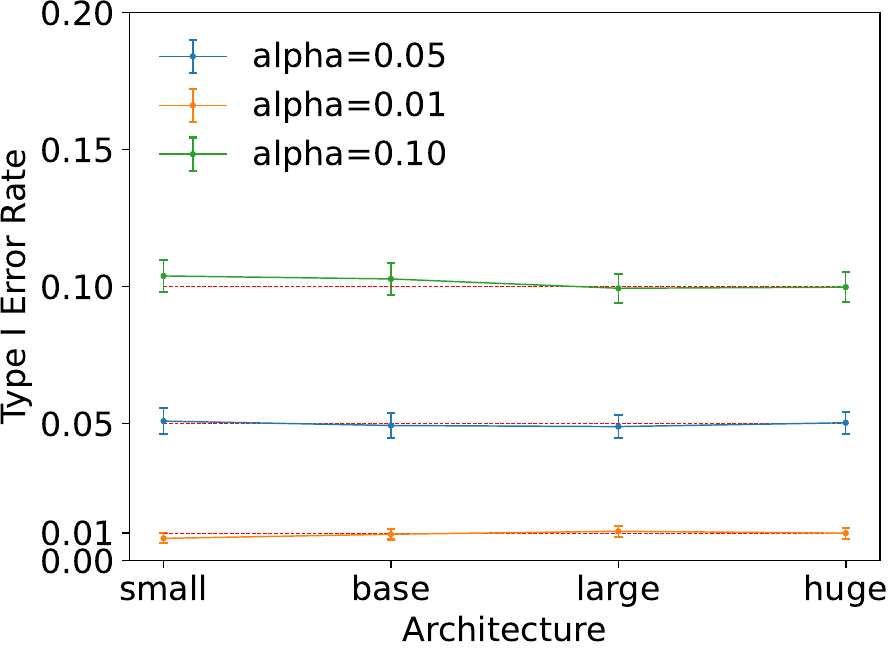}
        \subcaption{Architecture}
    \end{minipage}
    \caption{Robustness of type I error rate control for estimated variance}
    \label{fig:robustness_estimated}
\end{figure}

In the case where the variance is estimated from the same data, we considered the same two options as in type I error rate experiments in \S\ref{sec:sec4}: the image size and the architecture.
For each setting, we generated 100 null images $X=(X_1,\ldots,X_n)\sim\mathcal{N}(\bm{0},I)$ and tested the type I error rate for three significance levels $\alpha=0.05,0.01,0.10$.
We run 100 trials (i.e., 10,000 null images in total).
Here, we used the sample variance of the same data to inference.
The results are shown in Figure~\ref{fig:robustness_estimated} and our proposed method can properly control the type I error rate.
\subsection{Non-Gaussian Noise.}
\begin{figure}[htbp]
    \begin{minipage}[b]{0.49\linewidth}
        \centering
        \includegraphics[width=1.0\linewidth]{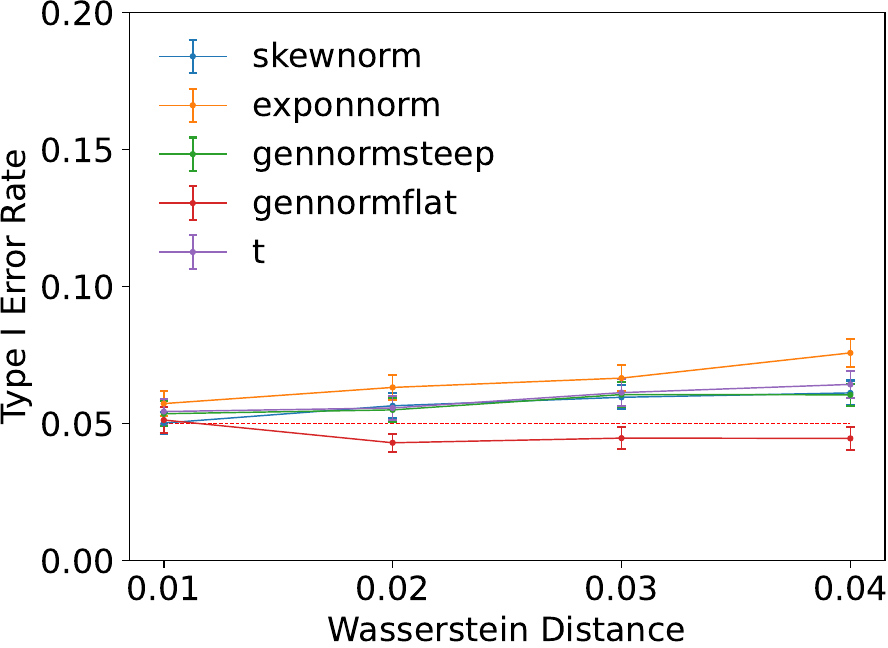}
        \subcaption{Significance Level: 0.05}
    \end{minipage}
    \begin{minipage}[b]{0.49\linewidth}
        \centering
        \includegraphics[width=1.0\linewidth]{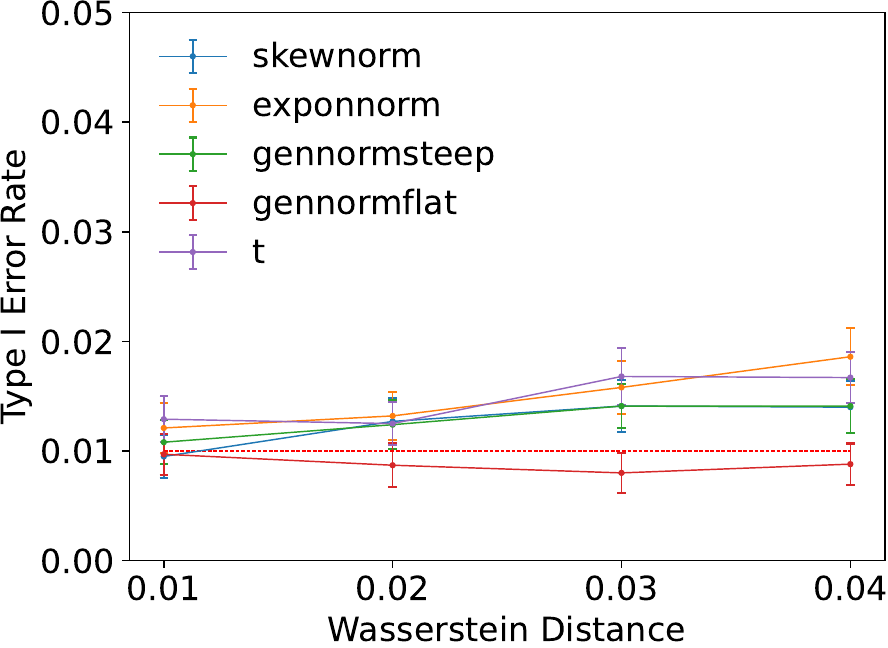}
        \subcaption{Significance Level: 0.01}
    \end{minipage}
    \caption{Robustness of type I error rate control for non-Gaussian noise}
    \label{fig:robustness_non_gaussian}
\end{figure}

In the case where the noise is non-Gaussian, we set the image size to 256 and the architecture to base.
As non-Gaussian noise, we considered the following five distribution families: \texttt{skewnorm}, \texttt{exponnorm}, \texttt{gennormsteep}, \texttt{gennormflat} and \texttt{t}.
The details of these distribution families are shown in Appendix~\ref{app:non_gaussian}.
To conduct the experiment, we first obtained a distribution such that the 1-Wasserstein distance from $\mathcal{N}(0,1)$ is $d$ in each distribution family, for $d\in\{0.01,0.02,0.03,0.04\}$.
We then generated 100 images following each distribution and tested the type I error rate for two significance levels $\alpha=0.05,0.01$.
We run 100 trials (i.e., 10,000 null images in total).
The results are shown in Figure~\ref{fig:robustness_non_gaussian}.
Only for \texttt{gennormflat}, the type I error rate is smaller than the significance level, i.e., the proposed method is conservative.
For other three distribution families \texttt{skewnorm}, \texttt{gennormsteep} and \texttt{t}, our proposed method still can properly control the type I error rate, albeit slightly higher than the significance level.
In case of \texttt{exponnorm}, our proposed method seem to fail to control the type I error rate.
\subsection{Details of Non-Gaussian Noise Distributions}
\label{app:non_gaussian}
\begin{figure}[htbp]
    \centering
    \includegraphics[width=0.6\linewidth]{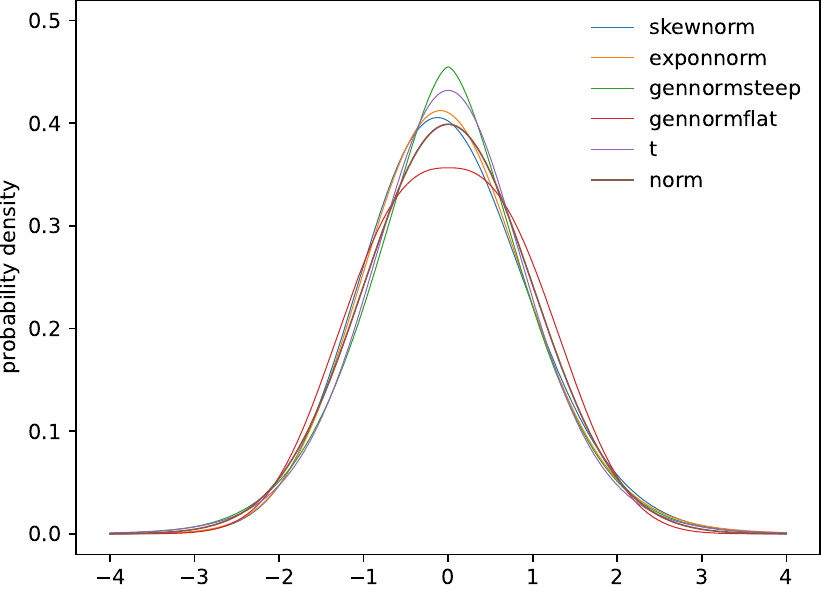}
    \caption{Demonstration of non-Gaussian distributions}
    \label{fig:non_gaussian_pdf}
\end{figure}

We considered the following five non-Gaussian distribution families:
\begin{itemize}
    \item \texttt{skewnorm}: Skew normal distribution family.
    \item \texttt{exponnorm}: Exponentially modified normal distribution family.
    \item \texttt{gennormsteep}: Generalized normal distribution family (limit the shape parameter $\beta$ to be steeper than the normal distribution, i.e., $\beta < 2$).
    \item \texttt{gennormflat}: Generalized normal distribution family (limit the shape parameter $\beta$ to be flatter than the normal distribution, i.e., $\beta > 2$).
    \item \texttt{t}: Student's t distribution family.
\end{itemize}
Note that all of these distribution families include the Gaussian distribution and are standardized in the experiment.
We demonstrate the probability density functions for distributions from each distribution family such that the 1-Wasserstein distance from $\mathcal{N}(0,1)$ is $0.04$ in Figure~\ref{fig:non_gaussian_pdf}

\newpage
\section{Additional Trials for Type I Error Rate}
\label{app:additional_trials}
\begin{figure}[htbp]
    \begin{minipage}[b]{0.49\linewidth}
        \centering
        \includegraphics[width=1.0\linewidth]{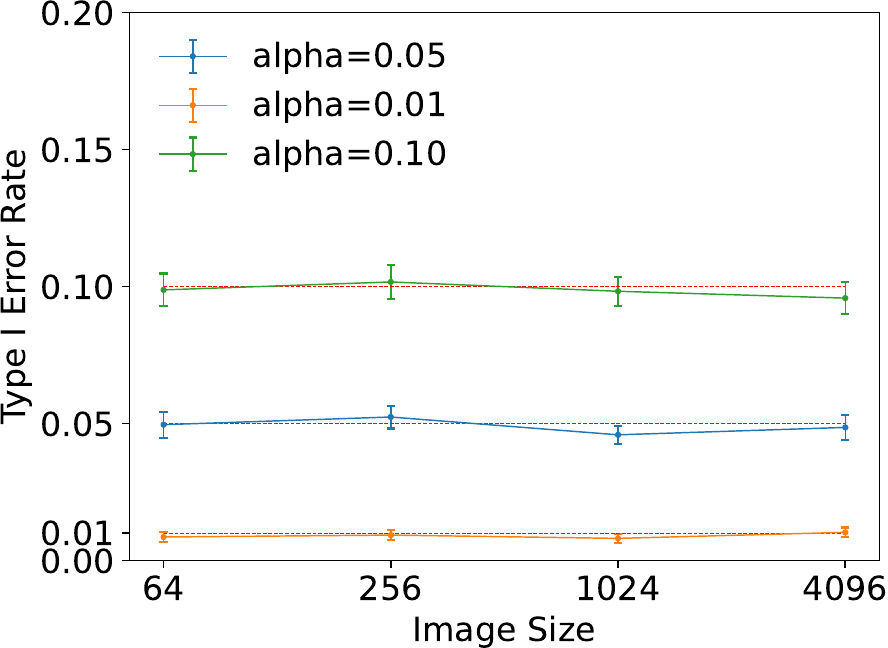}
        \subcaption{Independence}
    \end{minipage}
    \begin{minipage}[b]{0.49\linewidth}
        \centering
        \includegraphics[width=1.0\linewidth]{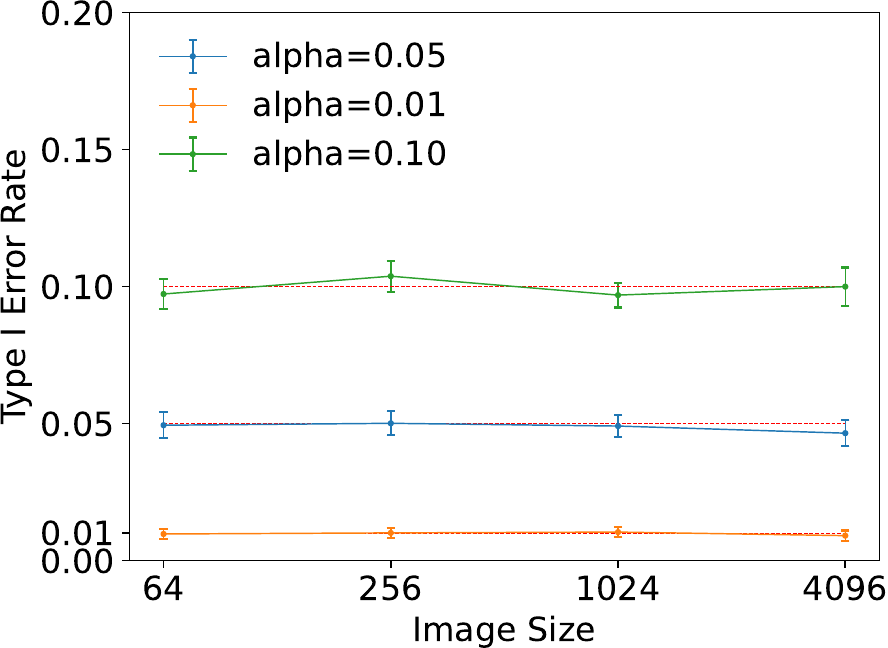}
        \subcaption{Correlation}
    \end{minipage}
    \caption{Type I Error Rate for 10,000 iterations (image size)}
    \label{fig:additional_fpr_image_size}
\end{figure}

\begin{figure}[htbp]
    \begin{minipage}[b]{0.49\linewidth}
        \centering
        \includegraphics[width=1.0\linewidth]{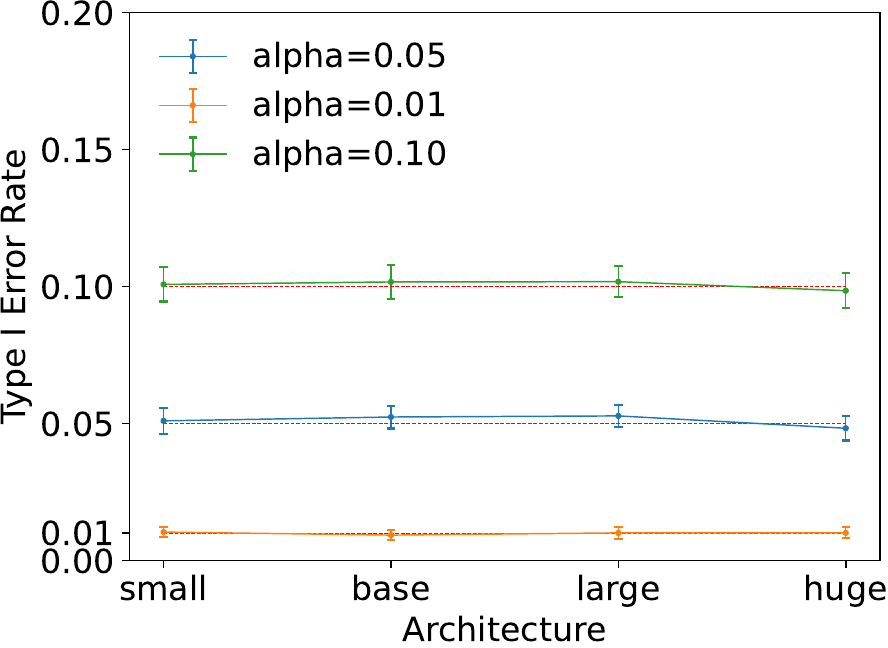}
        \subcaption{Independence}
    \end{minipage}
    \begin{minipage}[b]{0.49\linewidth}
        \centering
        \includegraphics[width=1.0\linewidth]{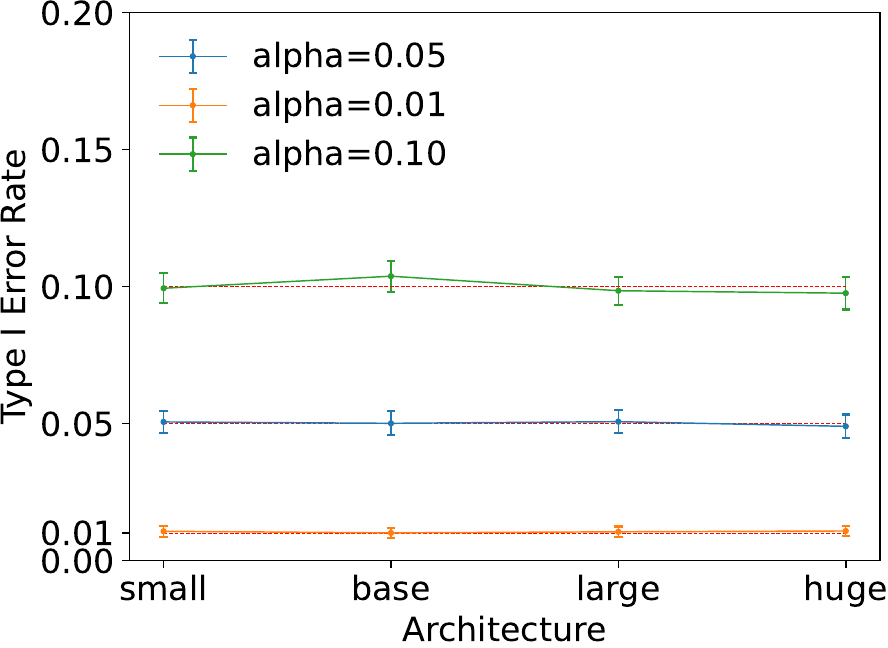}
        \subcaption{Correlation}
    \end{minipage}
    \caption{Type I Error Rate for 10,000 iterations (architecture)}
    \label{fig:additional_fpr_architecture}
\end{figure}

%
%
%
%
We already run the 10 trials of 100 null images for each setting and confirmed that our proposed method can properly control the type I error rate in \S\ref{sec:sec4}.
Here, we run additional 90 trials of 100 null images (i.e., 10,000 null images in total) for each setting and examined the validity of our proposed method more carefully.
We considered the three significance levels $\alpha=0.05,0.01,0.10$ in this experiment.
The results are shown in Figures~\ref{fig:additional_fpr_image_size} and \ref{fig:additional_fpr_architecture}.
We confirmed that our proposed method can properly control the type I error rate for many iterations.

\newpage
\bibliographystyle{plainnat}
\bibliography{ref}

\end{document}